\documentclass[lettersize,journal]{IEEEtran}
\usepackage{amsmath,amsfonts,amssymb}
\usepackage{algorithmic}
\usepackage{algorithm}
\usepackage{array}
\usepackage[caption=false,font=normalsize,labelfont=sf,textfont=sf]{subfig}
\usepackage{textcomp}
\usepackage{stfloats}
\usepackage{url}
\usepackage{verbatim}
\usepackage{graphicx}
\usepackage{cite}
\usepackage{siunitx}
\usepackage{subcaption}
\usepackage{multirow}
\usepackage{hyperref}

\begin{document}

\title{Progressive Experience Fusion for Multi-Task World Model Control in Endovascular Navigation}


\author{
Harry~Robertshaw$^{1}$, 
Maxence Boels$^{1}$, 
Nikola Fischer$^{1}$,
Sebastien Ourselin$^{1}$,
Christos Bergeles$^{1}$,
Alejandro~Granados$^{1}$
and~Thomas~C~Booth$^{1,2}$
\thanks{Partial financial support was received from the WELLCOME TRUST (Grant Agreement No 203148/A/16/Z), the Engineering and Physical Sciences Research Council Doctoral Training Partnership (Grant Agreement No EP/R513064/1), and the MRC IAA 2021 King's College London (MR/X502923/1). For the purpose of Open Access, the Author has applied a CC BY public copyright license to any Author Accepted Manuscript version arising from this submission.}
\thanks{Work in the Surgical \& Interventional Engineering Validation Suite (mock OR) has been supported by core funding from the Wellcome/EPSRC Centre for Medical Engineering [WT203148/Z/16/Z], a multi-user equipment grant for post-mortem evaluation of medical devices from Wellcome [218286/Z/19/Z], and the Wolfson Foundation [PR/ylr/md/21896].}%
\thanks{$^{1}$Harry Robertshaw, Maxence Boels, Nikola Fischer, Sebastien Ourselin, Christos Bergeles, Alejandro Granados, and Thomas C Booth are with Surgical \& Interventional Engineering, School of Biomedical Engineering \& Imaging Sciences, King's College London, UK}%
\thanks{$^{2}$Thomas C Booth is with the Department of Neuroradiology, King's College Hospital, UK {\tt\small thomas.booth@kcl.ac.uk}}%
\thanks{This work has been submitted to the IEEE for possible publication. Copyright may be transferred without notice, after which this version may no longer be accessible.}
}



\maketitle

\begin{abstract}

    Autonomous endovascular navigation could support the delivery of mechanical thrombectomy to underserved areas, but controllers must navigate long, multi-stage paths across varying vascular anatomies. This study investigates Progressive Experience Fusion (PEF) to train a multi-task TD-MPC2 controller. We additionally evaluate a heuristic that changes the Model Predictive Path Integral planning horizon using residual action-sequence dispersion, and fine-tuning in a patient-specific simulation. Across five subtasks in ten known training anatomies with held-out targets, PEF achieved a mean success rate of $74\%$, compared with $37\%$ for Soft Actor-Critic ($p < 0.001$) and 65\% for base TD-MPC2 ($p = 0.053$). A PEF controller with adaptive-horizon planning trained on 30 vasculatures achieved a mean success rate of $90$\% in ten held-out vasculatures. The PEF agent successfully transferred to an unseen \textit{in vitro} stroke patient vasculature under fluoroscopy, achieving a mean path ratio improvement from $63\%$ to $80\%$ with fine-tuning ($p < 0.001$), following $40 \times 10^3$ fine-tuning steps (corresponding to $\approx 107$\,\unit{\minute} of clinical inter-hospital transfer time). This work represents a proof of concept for multi-vasculature training and patient-specific adaptation, while further validation is required before clinical deployment.

\end{abstract}



\section{INTRODUCTION}

    \IEEEPARstart{T}{he} global burden of ischemic stroke continues to rise. In 2019, 77.19~million individuals experienced an ischemic stroke worldwide, with projections indicating a continued increase across all age groups by 2030~\cite{Feigin2021,Pu2023}. In the US alone, there is a substantial economic impact, with total costs projected to surge from \$67~billion in 2020 to \$423~billion by 2050, marking a $535\%$ increase~\cite{Kazi2024}. Mechanical thrombectomy (MT) has become the standard of care for acute ischemic stroke due to large vessel occlusion, offering improved functional outcomes when compared with medical treatment alone~\cite{Bendszus2023, Nogueira2018}. MT involves sequentially navigating multiple endovascular devices - including a guide catheter, access catheter, micro-catheter, and micro-guidewire - through tortuous vasculature to reach and remove a cerebral thrombus, typically located in the M1 segment of the middle cerebral artery (MCA). Procedure times average around 60\,\unit{\minute}~\cite{Weddell2020}.
    
    Timely intervention is critical, as while MT can be effective even up to $24$\,\unit{\hour} after symptom onset, the clinical benefit diminishes rapidly over time~\cite{Saver2016, Nogueira2018, Asdaghi2023}. Despite its proven efficacy, only $3–4\%$ of stroke admissions in, for example, the UK receive MT, though an estimated $15\%$ are eligible~\cite{SSNAP2024, McMeekin2024}. Barriers include geographic limitations, as many patients live outside MT-capable catchment areas, and significant inter-hospital transfer delays. In the UK, patients with door-to-door transfer times exceeding $3$\,\unit{\hour} (which occurs in $45\%$ of cases) are less likely to receive MT~\cite{Zhang2021}. Similarly, in the US, the probability of undergoing MT decreases by $1\%$ for every minute of transfer delay beyond the ideal $1$\,\unit{\hour} window~\cite{Regenhardt2018}.

    Moreover, occasional procedural complications such as vessel perforations ($1\%$), procedure-related vessel dissections ($2\%$), or distal embolization of thrombus ($9\%$)~\cite{Berkhemer2015}, combined with operator risks from prolonged radiation exposure, further highlight the need for innovation~\cite{Klein2009}. Additionally, protective gear designed to mitigate radiation exposure may itself increase orthopedic strain on operators~\cite{Madder2017}.

    Robotic surgical systems provide a promising solution by improving procedural access and reducing operator burden~\cite{Robertshaw2023}. Tele-operated MT from centralized neuroscience centers can enable expert neurointervention in underserved regions, while AI-assisted robotic systems can empower less experienced clinicians, such as general interventionalists (i.e., those performing endovascular procedures but not neurointerventional procedures related to ischemic or hemorrhagic stroke) in non-specialist hospitals to perform MT effectively. With only 27~neuroscience centers in the UK compared to hundreds of non-specialist hospitals~\cite{SSNAP2023}, there is a clear need for AI-assisted, scalable robotic solutions. The integration of AI into robotic MT systems has the potential to enhance procedural efficiency and safety, with studies demonstrating that autonomous surgical robots can match or surpass human performance in controlled settings, indicating a considerable advancement in the field~\cite{RiveroMoreno2024}.

    To our knowledge, autonomous endovascular navigation has not yet been evaluated \textit{in vivo} for MT. The majority of experiments have been conducted \textit{in silico} with some performed \textit{in vitro}. The latter typically have been performed on non-anatomical vessel platforms `idealized' for simple navigation, or smaller anatomical sections~\cite{Zhang2025,Yao2025}. Both these types of \textit{in vitro} experiments can be classified as `short' navigation tasks. However, reinforcement learning (RL) can struggle to adapt to complex tasks over longer periods~\cite{Yu2019}. MT would be classified as a `long' navigation task, as multiple difficult inputs over an extended period of time are required to navigate the complex human vasculature. 
    
    The feasibility of RL-based autonomous endovascular navigation has been demonstrated in the context of two MT navigation tasks: 1) navigating a guide catheter and guidewire from the femoral artery to the internal carotid artery (ICA)~\cite{Robertshaw2024}, and 2) subsequently navigating a micro-catheter and micro-guidewire from the ICA to the MCA on unseen patient vasculatures~\cite{Robertshaw1_2025}. Both these studies were limited to \textit{in silico} work, and in the first, a single vasculature was used for training and testing~\cite{Robertshaw2024}. Recently, Hierarchical Modular Multi-Agent RL was shown to transition this work to \textit{in vitro} testbeds, with the first \textit{in vitro} navigation of the MT vasculature~\cite{Robertshaw2026ral}. However, this study did not perform navigation tasks on unseen (hold-out) vascular environments. 

    \begin{figure*}[tb]
        \centering
        \includegraphics[width=0.95\linewidth]{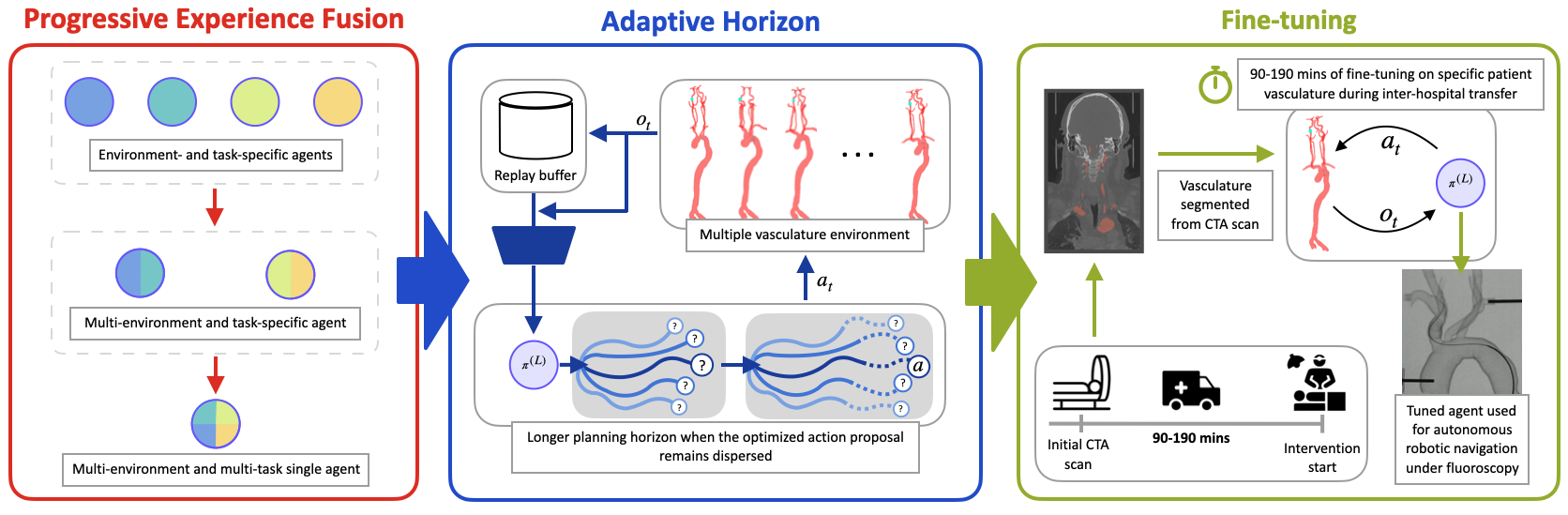}
        \caption{ Study overview. Progressive Experience Fusion (PEF) organizes replay data in two stages: experience is first grouped across vascular geometries within each navigation task and is then grouped across tasks to train one multi-task TD-MPC2 controller. At run time, TD-MPC2 evaluates candidate action sequences in its latent dynamics model using Model Predictive Path Integral (MPPI) planning. The adaptive horizon heuristic increases the planning horizon when the final MPPI action-sequence proposal remains dispersed. The generalist agent is fine-tuned \textit{in silico} using time-limited patient-specific interaction prior to intervention, enabling adaptation to previously unseen vascular anatomies without modifying hyperparameters.}
        \label{fig:overview}
    \end{figure*}

    RL algorithms are generally sensitive to architecture and hyperparameters, struggle to perform over long time horizons, and are often designed for single-task learning only~\cite{Henderson2018,Georgiev2024}. For complex or multiple tasks, this limits RL to computationally expensive models~\cite{Hafner2023}. To move towards realizing the benefits of autonomous MT navigation, it is necessary to develop policies capable of performing long navigation tasks that can generalize to multiple, unseen patient vasculatures.

    One promising technique to overcome these challenges is world models, a learned representation of the environment that predicts its dynamics, enabling a single agent to optimize actions in a virtual setting without relying solely on real-world data \cite{Ha2018}. Using this model, large-scale learning from diverse datasets could create AI-based navigation systems that can understand, predict, and adapt to real-world complexities~\cite{Hu2023}. 
    
    World models, such as DreamerV3~\cite{Hafner2023} and TD-MPC2~\cite{Hansen2024}, have been shown to outperform specialized methods across diverse benchmark tasks. Recent work has shown the effectiveness of TD-MPC2 in the context of autonomous MT navigation compared to the previous state-of-the-art for autonomous endovascular interventions (Soft Actor-Critic (SAC))~\cite{Karstensen2025,Moosa2025} -- demonstrating the potential of world models for completing long navigation tasks across diverse anatomies. However, no unseen vasculature was tested, and evaluation was limited to \textit{in silico} rather than \textit{in vitro} environments~\cite{Robertshaw3_2025}.

    The primary aim of this study was therefore to evaluate whether one world model controller could perform multiple autonomous endovascular navigation tasks across multiple vascular geometries and transfer to a previously unseen \textit{in vitro} patient vasculature under fluoroscopy. To achieve this long navigation task, we investigate Progressive Experience Fusion (PEF), a framework to train a single agent capable of performing multi-dimensional tasks, which we apply to endovascular navigation tasks both \textit{in silico} and \textit{in vitro}, as shown in Fig.~\ref{fig:overview}. PEF enables the transfer of replay buffers from task-specific agents into a unified agent. We leverage TD-MPC2 and extend this by introducing an adaptive Model Predictive Path Integral (MPPI) horizon based on residual action-proposal dispersion. We first tested our proposed method without any fine-tuning phase to show its applicability to a case where a patient arrives at a center with a general interventionalist able to perform AI-assisted MT. We also investigated a post-training fine-tuning phase in which the agent is granted limited interaction time with each patient-specific vasculature before evaluation. This mirrors most real-world clinical workflows, where patients typically undergo initial imaging at a referring center before being transferred for intervention at an MT-capable center (time from initial scan to groin puncture is $\approx 190$\,\unit{\minute}~\cite{Sun2013}).

    Our objectives are threefold: (1) to evaluate the navigation performance of our approach on an unseen \textit{in vitro} vasculature under fluoroscopy; (2) to assess the effectiveness of the adaptive horizon on agent performance; and (3) to assess the effects of the inclusion of a clinically-motivated training pipeline. The contributions of this work are as follows:
    
    \begin{itemize}
        \item To our knowledge, we present the first autonomous MT navigation experiment in a previously unseen patient-derived \textit{in vitro} vasculature.
        \item We introduce the PEF framework, a method for training single agents for solving multiple tasks in unseen environments.
        \item We propose a clinically-motivated, patient-specific fine-tuning stage and evaluate its effect on navigation performance in previously unseen anatomies.
        \item We propose an adaptive planning horizon that reduces residual distance to target in unsuccessful navigations.
        \item We benchmark the proposed PEF controller against SAC and base TD-MPC2 controllers previously evaluated for autonomous endovascular navigation.
    \end{itemize}

\section{METHODS}

    \subsection{Navigation tasks}

\begin{figure}[tb]
            \centering
            \includegraphics[width=0.75\linewidth]{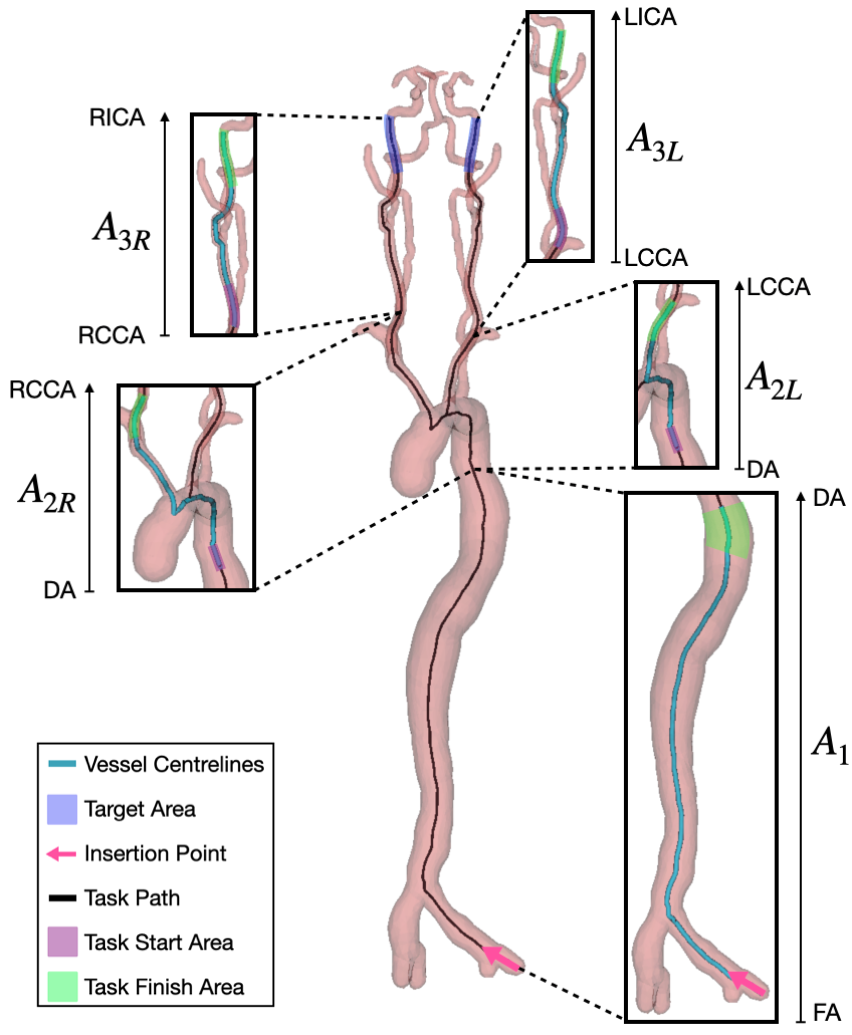}
            \caption{ Navigation tasks investigated. A target was randomly sampled from the set of points within the task finish area, while a guide catheter and guidewire navigated to it from a randomly sampled point in the task start area. DA: descending aorta, LCCA: left common carotid artery, LICA: left internal carotid artery, RCCA: right common carotid artery, RICA: right internal carotid artery.}
            \label{fig:tasks}
        \end{figure}

        The first `phase' of MT typically involves navigating a guide catheter from the femoral or radial artery to the ICA. An `access catheter' (with a shaped tip) is typically placed within the guide catheter and taken ahead of the guide catheter tip during navigation. The guide catheter is then advanced to make a stable platform for the second phase, where a micro-guidewire and micro-catheter may be navigated to the thrombus site to enable a stent retriever to remove the thrombus. The first phase of MT has been split into five subtasks for multi-task RL training, and is described in Fig.~\ref{fig:tasks}. 
        

    \subsection{Dataset}\label{dataset}

        To provide a suitable dataset for the training and testing of RL agents that can navigate unseen vascular environments \textit{in silico} and \textit{in vitro}, 40~patient vasculatures were used in this study. A `neck computed tomography angiography (CTA)' scan of a stroke patient that encompassed the aortic arch to the cerebral vessels, and a `body CTA' scan encompassing the abdominal and thoracic regions, including the femoral arteries, descending aorta, and the aortic arch (obtained with UK Research Ethics approval 24/LO/0057), were manually processed into surface meshes, and arterial centerlines and radii were extracted using 3D Slicer (v5.8.0)~\cite{Fedorov2012}. To fuse the neck and body CTAs, the centerlines and radii of the abdominal and thoracic regions were scaled so that the radius at the most superior aspect of the descending aorta would match the neck CTA aorta centerline. These two sets of centerlines were joined, and the radius at each centerline point was used to generate a surface mesh, which was loaded into the Simulation Open Framework Architecture (SOFA, v23.12)~\cite{Faure2012}. This process was repeated for 40 neck CTAs (which remained unscaled), with the same body CTA scaled to fit each one — the rationale being that we wished to maintain realism for the neck vasculature, which is particularly heterogeneous and highly challenging for navigation in contrast to the body vasculature. These 40~vascular environments were used to create two types of datasets to be examined in this study:

        \begin{itemize}
            \item \textbf{Train10--Test0 dataset}: We use the same 10~patient vasculatures for both training and testing as in previous work, to first show that our proposed changes to TD-MPC2 provide improvements over the current state-of-the-art~\cite{Robertshaw3_2025}. For training, five target locations in each branch were used, and testing was performed on five different targets per branch. `Test0' refers to no unseen vasculature.
            \item \textbf{Train30--Test10 dataset}: 20~randomly selected patient vasculatures were then added to the Train10--Test0 dataset, to create a training set of 30. An additional 10 unseen patient vasculatures were used for testing to evaluate generalization. Ten targets in each branch were utilized for evaluation to comprehensively assess performance across the entire branch.
        \end{itemize}

    \subsection{Testbed}

        \subsubsection{In silico}

            The open-source stEVE framework was used to build an \textit{in silico} environment for the navigation task~\cite{Karstensen2025}. This utilizes the BeamAdapter plugin for SOFA, which was used to model the vasculatures and the devices used for navigation~\cite{Wei2012}. The devices modeled were a $0.035"$ guidewire (Terumo, Tokyo, Japan; $120$~vertices with Young's modulus of $43$\,\unit{\mega\pascal}) and a $0.0441"$ multipurpose (MP) catheter (Terumo, Tokyo, Japan; $160$~vertices with Young's modulus of $47$\,\unit{\mega\pascal})~\cite{Jackson2023}. 

            Feedback during the navigation was given as 2D tracking coordinates of three points along the device's tip equally spaced $2$\,\unit{\milli\meter} apart, denoted as $\mathbf{p}_t =[x'_{t,1},y'_{t,1},\ldots,x'_{t,3},y'_{t,3}]$, with $x'_{t,1},y'_{t,1}$ representing the instrument tip. No visual information, including the geometry of the patient vasculature, was available as observations during the navigation task. The target location, $p_{\mathrm{goal}}=(x_g',y_g')$, was resampled at the beginning of each episode and remained fixed throughout that episode. At time step $t$, the observation comprised the current and previous device-position vectors, $p_t$ and $p_{t-1}$, the target location, and the previous action, as defined in~\eqref{eq:obs}.

            \begin{equation}
                \mathbf{o}_t =
                [\mathbf{p}_t,\,
                 \mathbf{p}_{t-1},\,
                 \mathbf{p}_{\mathrm{goal}},\,
                 \mathbf{a}_{t-1}]
                \label{eq:obs}
            \end{equation}

            The action vector contained the translation and rotation speeds of both the guidewire and catheter. For each device, the translational velocities were bounded to $[-40,40]$\,\unit{\mm\per\second}, and rotational velocities were bounded to $[-180,180]$\,\unit{\degree\per\second}.

            A single-tracking (guidewire tracking only) method was employed, similar to \cite{Robertshaw2024}. The simulation's fidelity to real-world device behavior was achieved through iterative fine-tuning of parameters such as the friction between the device and vessel wall and the device's stiffness. Such methodology has previously allowed domain adaptation of autonomous navigation agents from \textit{in silico} to \textit{in vitro} models and from \textit{in silico} to \textit{ex vivo} of porcine liver vasculatures~\cite{Karstensen2025,Karstensen2022,Robertshaw2026ral}.

        \subsubsection{In vitro}

\begin{figure}[tb]
                \centering
                \includegraphics[width=0.7\linewidth]{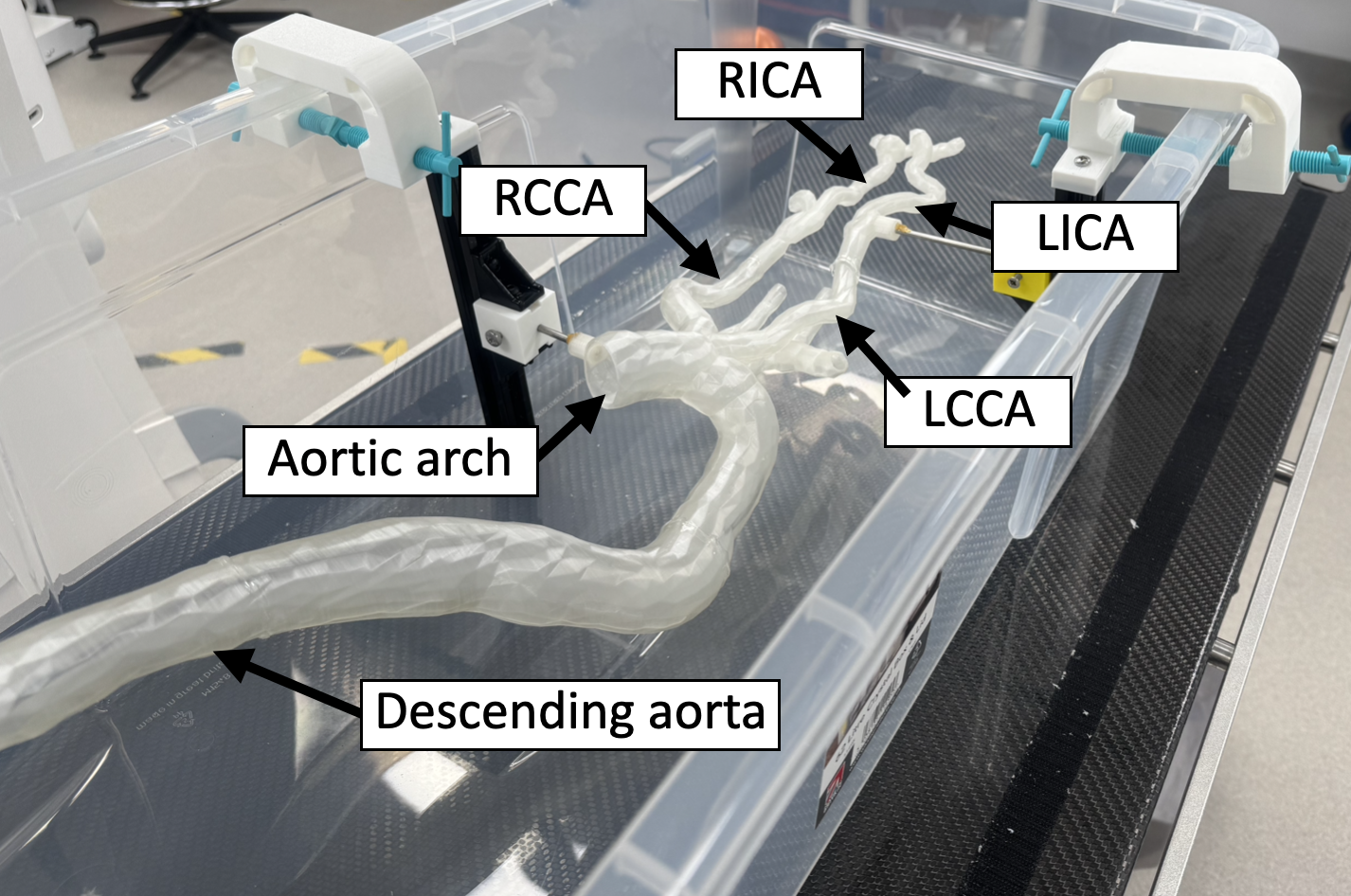}
                \caption{\textit{In vitro} 3D vascular phantom with labeled anatomy.}
                \label{fig:phantom}
            \end{figure}

            Our \textit{in vitro} testbed consisted of a custom 3D-printed transparent phantom (Clear V4, Formlabs Inc., Somerville, USA), as shown in Fig.~\ref{fig:phantom}. Two vascular phantoms were printed: one geometry from Train10--Test0 and one held-out geometry from Train30--Test10. Experiments were conducted in a mock operating room. A mobile C-arm (Cios Spin, Siemens Healthineers, Erlangen, Germany) was used to provide fluoroscopic images as an input to a device tip-tracking algorithm that extracted the necessary coordinates for the state-based RL algorithm~\cite{Eyberg2022}. A $180$\,\unit{\centi\metre} $0.035"$~guidewire (Terumo, Tokyo, Japan) and a $100$\,\unit{\centi\metre} $0.0441"$~MP catheter (Terumo, Tokyo, Japan) were used for experiments.

\begin{figure}[tb]
                \centering
                \includegraphics[width=0.7\linewidth]{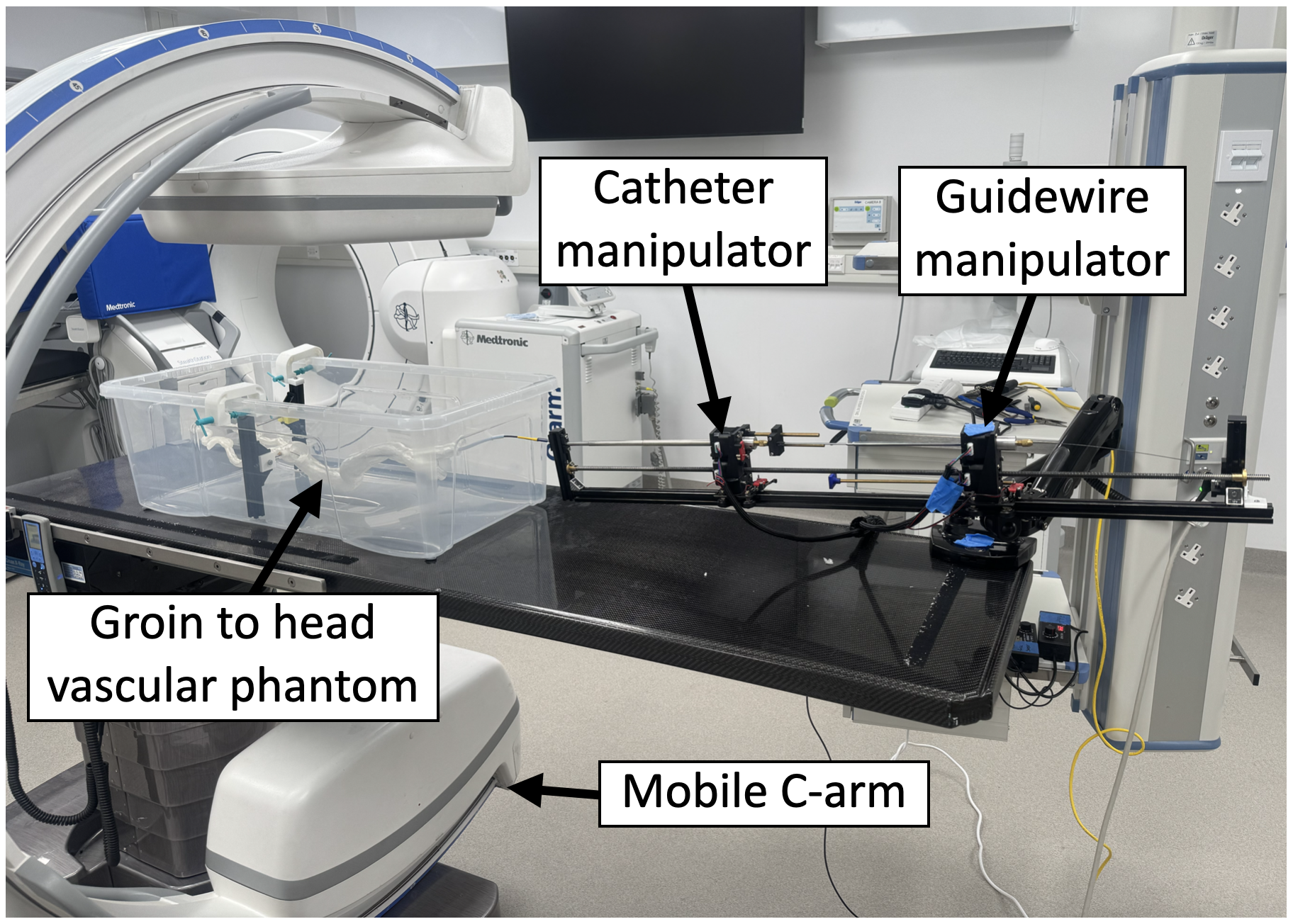}
                \caption{\textit{In vitro} testbed showing the 3D vascular phantom, mobile C-arm, and robotic manipulator.}
                \label{fig:in_vitro}
            \end{figure}

            A robotic manipulator (Fig.~\ref{fig:in_vitro}) was used to translate and rotate the catheter and guidewire independently~\cite{Sadati2025}. The system features a modular design with independent actuation carriage units, each capable of two degrees of freedom (rotation and translation), placed on a single lead screw rod, which was fixed on a frame made from aluminum struts. Direct-drive hollow shaft stepper motors were used with a leadscrew nut to translate the actuation carriage unit with an instrument along the length of the actuation system. Furthermore, the same stepper motor type was deployed to rotate the instruments via small drill chuck couplings. Telescopic tubes were installed between the carriages to prevent lateral buckling of the catheters and guidewires. The motors were controlled via G-code protocol using a standard CNC control board. 

    \subsection{Controller architecture}

        \subsubsection{Reward function}

        The dense reward function used during training across all models is shown in~\eqref{eq:R}~\cite{Robertshaw2026ral}. Here, \textit{pathlength} is defined as the distance between the guidewire tip and the target along the centerlines of the arteries, with $\Delta\text{pathlength}$ representing the change in pathlength at time $t$ from the previous step at time $t-1$.

        \begin{equation}
            r_t = -0.00015 - 0.001\cdot\Delta\text{pathlength}+\begin{cases}1 & \text{if target reached} \\0 & \text{else} \end{cases}
            \label{eq:R}
        \end{equation}   

        \subsubsection{Soft actor-critic}

            An implementation of the SAC RL algorithm from the stEVE framework was used to provide a baseline for comparison~\cite{Karstensen2025}. SAC is a model-free RL method that learns a stochastic policy to maximize expected reward while encouraging exploration through entropy regularization. This particular SAC implementation has been successfully used across several autonomous endovascular navigation studies~\cite{Karstensen2025,Robertshaw2024,Robertshaw1_2025}. We follow the implementation proposed in~\cite{Karstensen2022}, which includes a Long Short-Term Memory (LSTM) layer for learning trajectory-dependent state representations and feedforward layers for controlling the devices. The controller takes observations as input, and a Gaussian policy network outputs mean and standard deviation for expected actions, representing the catheter and guidewire rotations and translations. During training, actions are sampled from the mean and standard deviation, but for evaluation, mean is used directly for deterministic behavior.

        \subsubsection{TD-MPC2}\label{sec:tdmpc2}

            TD-MPC2 was used as an additional baseline and as the learning framework for the proposed agents~\cite{Hansen2024}. TD-MPC2 learns a compact latent world model comprising an observation encoder, latent dynamics model, reward predictor, ensemble of action-value functions, and stochastic policy prior. At each control step, the current observation is encoded into a latent state, after which candidate action sequences are evaluated by predicting their latent trajectories, rewards, and terminal values.

            Online control is performed using TD-MPC2’s MPPI-style trajectory optimizer~\cite{williams2015}. For the current planning horizon $H_t$, MPPI samples $N$ candidate action sequences from a factorized Gaussian proposal defined by a mean $\mu_{t,h}^{(j)}$ and standard deviation $\sigma_{t,h}^{(j)}$ at each horizon step $h$ and optimization iteration $j$. Of the $N$ candidate sequences, $N_\pi$ are generated by rolling out the learned policy prior, while the remaining $N-N_\pi$ are sampled from the Gaussian proposal. All sampled actions are bounded to the normalized action range. Each candidate sequence is propagated through the latent dynamics model and assigned the predicted return given by~\eqref{eq:mppi_return}.      

            \begin{equation}
                \begin{aligned}
                    G_t^{(n,j)}
                    ={}&
                    \sum_{h=0}^{H_t-1}
                    \gamma^h
                    r_\eta\!\left(
                    \hat{z}_{t,h}^{(n,j)},
                    a_{t,h}^{(n,j)}
                    \right) \\
                    &+
                    \gamma^{H_t}
                    \bar{Q}_\psi\!\left(
                    \hat{z}_{t,H_t}^{(n,j)},
                    \tilde{a}_{t,H_t}^{(n,j)}
                    \right), \\
                    \tilde{a}_{t,H_t}^{(n,j)}
                    \sim{}&
                    \pi_\xi\!\left(
                    \cdot \mid
                    \hat{z}_{t,H_t}^{(n,j)}
                    \right).
                \end{aligned}
                \label{eq:mppi_return}
            \end{equation}

            Here, $\gamma$ is the discount factor, $r_\eta$ is the learned reward predictor, and $\bar{Q}_\psi$ is the average of two randomly selected Q-functions. The terminal action is sampled from the learned policy prior, and the terminal Q-value estimates the return beyond the finite planning horizon. The $E$ candidates with the largest predicted returns form the elite set $\mathcal{E}_t^{(j)}$. Their returns are converted into normalized exponential weights as in~\eqref{eq:mppi_weights}.

            \begin{equation}
                \begin{aligned}
                    s_t^{(n,j)}
                    &=
                    \exp\!\left\{
                    \beta\!\left[
                    G_t^{(n,j)}-G_{t,\max}^{(j)}
                    \right]\right\},\\
                    w_t^{(n,j)}
                    &=
                    \frac{s_t^{(n,j)}}
                    {\displaystyle
                    \sum_{m\in\mathcal{E}_t^{(j)}}s_t^{(m,j)}},
                    \qquad
                    n\in\mathcal{E}_t^{(j)}.
                \end{aligned}
                \label{eq:mppi_weights}
            \end{equation}

            In~\eqref{eq:mppi_weights}, $G_{t,\max}^{(j)}$ is the largest elite return and $\beta$ controls how strongly the proposal is concentrated around the highest-return sequences. The weighted elite actions are used to update both the proposal mean and standard deviation according to~\eqref{eq:mppi_update}.

            \begin{equation}
                \begin{aligned}
                    \mu_{t,h}^{(j+1)}
                    &=
                    \sum_{n\in\mathcal{E}_t^{(j)}}
                    w_t^{(n,j)}a_{t,h}^{(n,j)},\\
                    v_{t,h}^{(j+1)}
                    &=
                    \sum_{n\in\mathcal{E}_t^{(j)}}
                    w_t^{(n,j)}
                    \left(
                    a_{t,h}^{(n,j)}
                    -\mu_{t,h}^{(j+1)}
                    \right)^2,\\
                    \sigma_{t,h}^{(j+1)}
                    &=
                    \min\!\left(
                    \sigma_{\max},
                    \max\!\left(
                    \sigma_{\min},
                    \sqrt{v_{t,h}^{(j+1)}}
                    \right)
                    \right).
                \end{aligned}
            \label{eq:mppi_update}
            \end{equation}

            The evaluation, weighting, and proposal update in~\eqref{eq:mppi_return}--\eqref{eq:mppi_update} are repeated for $K$ optimization iterations. As high-return action sequences become concentrated, the proposal standard deviation \eqref{eq:mppi_update} decreases, while a larger final standard deviation indicates that the high-return action sequences remain more dispersed. After the final iteration, an elite sequence is selected according to the weights in~\eqref{eq:mppi_weights}. Only the first action is applied to the environment, and planning is repeated after receiving the next observation. The optimized proposal mean is shifted forward by one step to initialize the following planning cycle.

            We used the $5\times10^6$-parameter TD-MPC2 configuration, while increasing the replay-buffer capacity from $1\times10^6$ to $1\times10^7$ transitions and the batch size from $256$ to $1024$. An LSTM-based observation embedder was additionally used to integrate sequential device-tracking observations, allowing its recurrent hidden state to retain information from the preceding device-tip trajectory.

        \subsubsection{Adaptive horizon}\label{sec:adaptive_horizon}

            Standard TD-MPC2 uses a fixed planning horizon, denoted $H_{\mathrm{base}}$. In the proposed adaptive-horizon method, the planning horizon used at each control step is recalculated relative to $H_{\mathrm{base}}$ using the residual dispersion of the MPPI action-sequence proposal. After the $K$ MPPI optimization iterations described in Section~\ref{sec:tdmpc2}, \eqref{eq:mppi_update} provides the final proposal standard deviation for every horizon step and action dimension. At control step $t$, these values are aggregated over the current planning horizon $H_t$ to obtain the mean residual dispersion defined by~\eqref{eq:sigma}.
            
            \begin{equation}
                \bar{\sigma}_t
                =
                \frac{1}{H_t d_a}
                \sum_{h=0}^{H_t-1}
                \sum_{q=1}^{d_a}
                \sigma_{t,h,q}^{(K)} 
                \label{eq:sigma}
            \end{equation}
            
            Here, $\sigma_{t,h,q}^{(K)}$ is the standard deviation of the final MPPI proposal at control step $t$, horizon step $h$, and action dimension $q$. Therefore, $\bar{\sigma}_t$ measures the residual dispersion of the optimized action proposal, rather than the variance of predicted latent states or a direct estimate of uncertainty in the learned dynamics. A smaller value indicates that the high-return candidate sequences have become concentrated around a narrower action proposal, whereas a larger value indicates that they remain more widely dispersed.
            
            The non-negative horizon extension relative to $H_{\mathrm{base}}$, denoted $\Delta H_t$, and the horizon used at the next control step are defined by~\eqref{eq:H}. Here, $\alpha>0$ controls the magnitude of the horizon extension, $H_{\max}$ is the maximum permitted integer horizon, and $\tau=0.2$ is the residual-dispersion threshold. At the beginning of each episode, the planning horizon is initialized as $H_0=H_{\mathrm{base}}=3$.
            
            \begin{equation}
                \begin{aligned}
                    \Delta H_t
                    &=
                    \left\lceil
                    \alpha
                    \left[
                        \frac{\bar{\sigma}_t}{\tau}-1
                    \right]_{+}
                    \right\rceil,\\
                    H_{t+1}
                    &=
                    \min\!\left(
                        H_{\max},
                        H_{\mathrm{base}}+\Delta H_t
                    \right).
                \end{aligned}
                \label{eq:H}
            \end{equation}
            
            Here, $[x]_{+}=\max(0,x)$ denotes the positive part of $x$. When $\bar{\sigma}_t\leq\tau$, $\Delta H_t=0$ and the planning horizon used at the next control step is reset to $H_{\mathrm{base}}$. When $\bar{\sigma}_t>\tau$, the horizon is extended relative to $H_{\mathrm{base}}$ in proportion to the threshold exceedance and is bounded by $H_{\max}$. A dispersed final proposal therefore increases the temporal coverage of the subsequent MPPI search, but does not imply that longer latent predictions are more accurate. The adaptive signal is interpreted as a measure of MPPI proposal concentration, not as a direct measure of model uncertainty or procedural safety.

    \subsection{Progressive Experience Fusion}

        While multi-task learning and hierarchical replay have been explored in previous work~\cite{Yin2017,Madden2004}, no prior work has described a framework of incremental merging replay buffers and subsequent retraining of hierarchical world models for endovascular navigation. Hierarchical Prioritized Experience Replay~\cite{Yin2017} and early Progressive Reinforcement Learning~\cite{Madden2004} introduced the idea of transferring experience between tasks. However, these approaches do not explicitly fuse full replay buffers and retrain successive world models. Our proposed PEF framework builds on these principles by performing explicit, layer-wise replay buffer fusion and model retraining with the aim of progressively generalizing across multiple dimensions of related tasks. It is also possible to combine the PEF framework with an adaptive horizon during agent training.

        Multi-task TD-MPC2 world models have previously been trained on a dataset of the replay buffers of single-task agents, resulting in a single TD-MPC2-trained agent able to complete a variety of tasks~\cite{Hansen2024}. Other work utilized this in combination with SAC-trained single-task agents to create an agent capable of performing multiple endovascular navigation tasks across a known dataset~\cite{Robertshaw3_2025}. PEF extends this concept by introducing a hierarchical training framework for incremental generalization across anatomies and tasks (Fig.~\ref{fig:pef_general_pyramid}). This is done by constructing a pyramid of agents that are progressively fused across multiple layers, transferring knowledge upward from task-specific to generalist representations. 

        \subsubsection{Theory}

\begin{figure}[tb]
                \centering
                \includegraphics[width=1\linewidth]{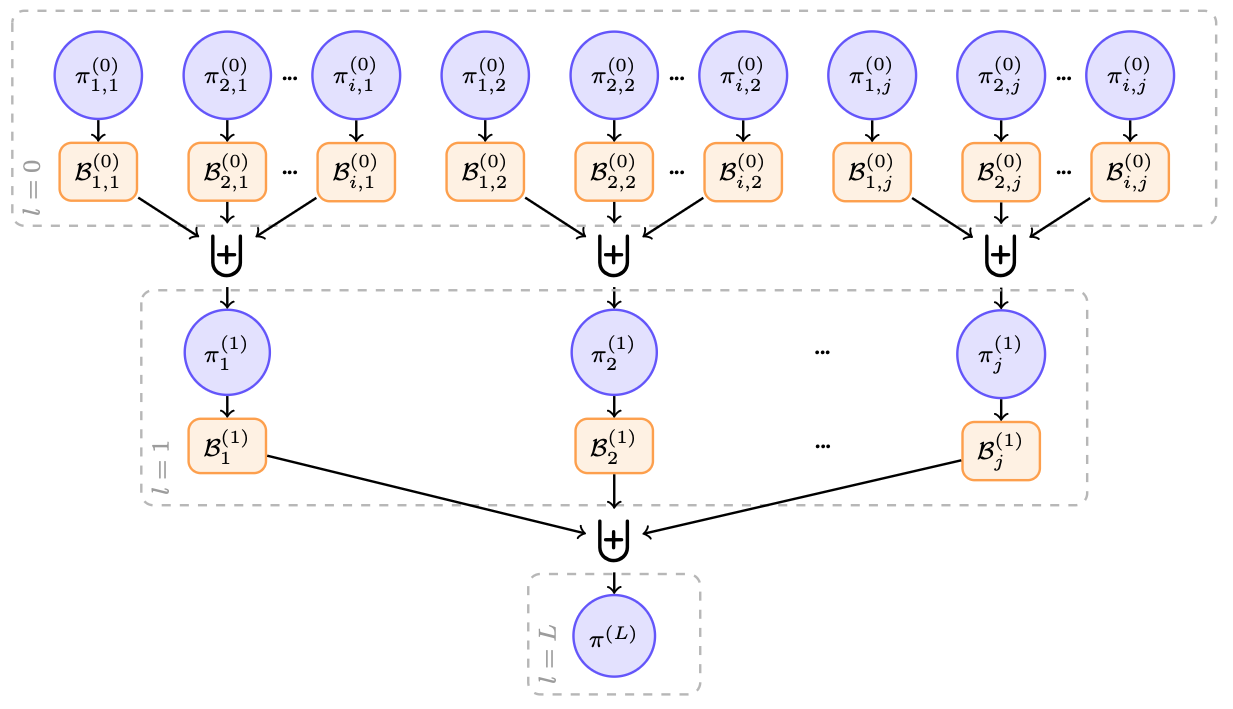}
                \caption{Illustration of the Progressive Experience Fusion (PEF) process.
                Each agent at layer $l>0$ is initialized from the fused post-training replay buffers of its assigned agents at layer $l-1$ and is subsequently trained across the combined configuration set. Its updated post-training replay buffer is then passed to the next fusion layer, culminating in the generalist policy $\pi^{(L)}$.}
                \label{fig:pef_general_pyramid}
            \end{figure}

            PEF organizes the task space into \(D\) dimensions, each representing a factor of variation (e.g., navigation subtask, environment). One fusion layer is applied per dimension (Fig.~\ref{fig:pef_general_pyramid}), such that experience is first combined within one factor and subsequently across the remaining factors. The total number of dimensions defines the structure and height of the fusion pyramid.
            
            At the base layer ($l=0$), each agent is trained independently on one task configuration. Let $\mathcal{C}_{\mathbf{i}}^{(0)}=\{\mathbf{i}\}$ denote the configuration assigned to base agent $\mathbf{i}\in\mathcal{I}$. We use $\operatorname{Train}(\mathcal{C};\mathcal{B}_{\mathrm{init}})$ to denote TD-MPC2 training across the configuration set $\mathcal{C}$ with replay buffer $\mathcal{B}_{\mathrm{init}}$. The operation returns the trained policy and its post-training replay buffer. The base agents and buffers are therefore obtained using~\eqref{eq:pef_base_training}.

            \begin{equation}
                \left(
                \pi_{\mathbf{i}}^{(0)},
                \mathcal{B}_{\mathbf{i}}^{(0)}
                \right)
                =
                \operatorname{Train}\!\left(
                \mathcal{C}_{\mathbf{i}}^{(0)};
                \varnothing
                \right),
                \qquad
                \mathbf{i}\in\mathcal{I}.
                \label{eq:pef_base_training}
            \end{equation}

            Here, $\mathcal{B}_{\mathbf{i}}^{(0)}$ contains the transitions collected by the individual agent during training, with each transition taking the form $(o_{e,t},a_{e,t},r_{e,t},o_{e,t+1})$, where $e$ and $t$ index the episode and time step, respectively. At each subsequent fusion layer $l>0$, agents from layer $l-1$ are partitioned into groups $\{\mathcal{S}_k^{(l)}\}_{k=1}^{M_l}$. For each group, the represented task configurations are combined and the post-training replay buffers from the preceding layer are fused as in~\eqref{eq:pef_layer_inputs}.

            \begin{equation}
            \begin{aligned}
                \mathcal{C}_k^{(l)}
                &=
                \bigcup_{m\in\mathcal{S}_k^{(l)}}
                \mathcal{C}_m^{(l-1)},                                      \\[1mm]
                \widetilde{\mathcal{B}}_k^{(l)}
                &=
                \biguplus_{m\in\mathcal{S}_k^{(l)}}
                \mathcal{B}_m^{(l-1)} .
            \end{aligned}
            \label{eq:pef_layer_inputs}
            \end{equation}

            Here, $\widetilde{\mathcal{B}}_k^{(l)}$ is the fused replay buffer used to initialize the higher-level agent. A new agent is then trained across the broader configuration set $\mathcal{C}_k^{(l)}$, as defined in~\eqref{eq:pef_layer_training}.

            \begin{equation}
                \left(
                \pi_k^{(l)},
                \mathcal{B}_k^{(l)}
                \right)
                =
                \operatorname{Train}\!\left(
                \mathcal{C}_k^{(l)};
                \widetilde{\mathcal{B}}_k^{(l)}
                \right)
                \label{eq:pef_layer_training}
            \end{equation}

            During this training operation, the replay buffer is initialized with $\widetilde{\mathcal{B}}_k^{(l)}$ and subsequently updated with transitions generated by the new higher-level agent. Consequently, $\mathcal{B}_k^{(l)}$ denotes the post-training buffer passed to the next fusion layer, rather than merely the unchanged union of the preceding buffers. The process is repeated until $M_L=1$, leaving the final generalist policy $\pi^{(L)}=\pi_1^{(L)}$ and its replay buffer $\mathcal{B}^{(L)}=\mathcal{B}_1^{(L)}$.

        \subsubsection{Implementation}

            In our implementation, PEF has $D=L=2$, with environment and navigation subtask forming the two dimensions. For $N_{\mathrm{env}}$ environments and $N_{\mathrm{task}}=5$ subtasks, the base layer contains $M_0=N_{\mathrm{env}}N_{\mathrm{task}}$ single-task agents. Each base agent $\pi_{i,j}^{(0)}$ is trained independently on environment $i$ and subtask $j$, producing the post-training replay buffer $\mathcal{B}_{i,j}^{(0)}$.

            At the first fusion layer ($l=1$), the post-training buffers of base agents performing the same subtask are combined across environments. The resulting buffer $\widetilde{\mathcal{B}}_j^{(1)}$ is used to initialise the replay buffer of a new task-specialist agent. This agent is trained across the configuration set $\mathcal{C}_j^{(1)}$, which contains subtask $j$ in all $N_{\mathrm{env}}$ environments. The fusion and subsequent training operations are defined in~\eqref{eq:MT_B1}.

            \begin{equation}
            \begin{aligned}
                \widetilde{\mathcal{B}}_j^{(1)}
                &=
                \biguplus_{i=1}^{N_{\mathrm{env}}}
                \mathcal{B}_{i,j}^{(0)},                                    \\[1mm]
                \left(
                \pi_j^{(1)},
                \mathcal{B}_j^{(1)}
                \right)
                &=
                \operatorname{Train}\!\left(
                \mathcal{C}_j^{(1)};
                \widetilde{\mathcal{B}}_j^{(1)}
                \right),
            \end{aligned}
            \qquad
            j=1,\ldots,N_{\mathrm{task}} .
            \label{eq:MT_B1}
            \end{equation}

            Here, $\mathcal{B}_j^{(1)}$ is the post-training replay buffer of the task-specialist agent. It is initialized using $\widetilde{\mathcal{B}}_j^{(1)}$ and updated with transitions generated during task-specialist training. Fusion across the environment dimension produces $M_1=N_{\mathrm{task}}=5$ trained task-specialist agents and their associated buffers.

            At the second fusion layer ($l=2$), the post-training buffers of the task-specialist agents are combined across subtasks. The resulting buffer $\widetilde{\mathcal{B}}^{(2)}$ initializes the replay buffer used to train the final generalist agent across the complete configuration set $\mathcal{C}^{(2)}$. These operations are defined in~\eqref{eq:MT_B2}.

            \begin{equation}
            \begin{aligned}
                \widetilde{\mathcal{B}}^{(2)}
                &=
                \biguplus_{j=1}^{N_{\mathrm{task}}}
                \mathcal{B}_j^{(1)},                                        \\[1mm]
                \left(
                \pi^{(2)},
                \mathcal{B}^{(2)}
                \right)
                &=
                \operatorname{Train}\!\left(
                \mathcal{C}^{(2)};
                \widetilde{\mathcal{B}}^{(2)}
                \right).
            \end{aligned}
            \label{eq:MT_B2}
            \end{equation}

            Fusion across the subtask dimension therefore produces $M_2=1$ generalist agent that performs all navigation subtasks across the corresponding vascular-anatomy distribution. The layer sizes are $(M_0,M_1,M_2)=(50,5,1)$ for Train10--Test0 ($N_{\mathrm{env}}=10$) and $(150,5,1)$ for Train30--Test10 ($N_{\mathrm{env}}=30$). Counting all layers, these configurations contain $56$ and $156$ trained agent--buffer nodes, respectively.

    \subsection{Fine-tuning}

        We aimed to leverage both the adaptability of world model algorithms, such as TD-MPC2, and the current MT treatment pathway to maximize performance on previously unseen vasculatures. In practice, most patients initially present at non-MT-capable centers where CTA images are obtained and sent to MT-capable centers by teleradiology. If MT is approved by the MT-capable center, the patient will undergo inter-hospital patient transfer for definitive treatment. 
        
        To make use of this redundant transfer time, a patient-specific simulator was constructed from the pre-operative CTA, and a copy of the pretrained generalist agent underwent time-bounded continued training through simulated interaction before evaluation. No hyperparameters were changed during the fine-tuning process. The reported fine-tuning times include controller training only and exclude CTA segmentation, mesh generation, simulation construction and data-transfer time.

    \subsection{Training, experimental design and evaluation}

        \subsubsection{Training}

            Each agent was trained for $1\times10^6$ environment steps (or until it reached $100\%$ success rate) using the same TD-MPC2 parameters. All training was completed on an NVIDIA DGX A100 node (Santa Clara, California, USA). Each navigation task was considered an `episode' and was deemed complete once the target was reached within a pre-specified target boundary defined as $\pm$ distance equivalent to vessel diameter. To ensure computational efficiency, we introduced a training timeout after $200$~exploration steps (equivalent to approximately 27\,s) without reaching the target.

            Across both datasets, vasculatures in the training set were augmented using a scaling function for height and width, which was implemented to increase generalisability. A scaling factor between 0.7 and 1.3 for height and width was randomly assigned to the vasculature before each navigation attempt. Augmentation was not applied to the test set.

        \subsubsection{Experimental design}

            A series of experiments were conducted to evaluate the proposed PEF framework across both \textit{in silico} and \textit{in vitro} environments.
            
            \begin{enumerate}
                \item \textbf{\textit{In silico} validation on the Train10--Test0 dataset:} The first experiment replicated the \textit{in silico} navigation tasks from the previous state-of-the-art~\cite{Robertshaw3_2025}, using the Train10--Test0 dataset to validate improvements introduced by the PEF framework.
                \item \textbf{\textit{In vitro} validation on a Train10--Test0 vasculature:} Agents trained on the Train10--Test0 dataset were applied to an \textit{in vitro} vascular phantom experiment (selected from the Train10--Test0 dataset), to assess whether PEF-trained policies can successfully transfer from simulation to a physical environment, while comparing results to previous baselines~\cite{robertshaw2026iros}.
                \item \textbf{\textit{In silico} generalization on the Train30--Test10 dataset:} A PEF agent was trained on 30, and tested on 10 unseen vasculatures. The effects of the adaptive horizon on navigation performance were also analyzed.
                \item \textbf{Fine-tuning on unseen \textit{in silico} vasculatures:} A fine-tuning study was performed to quantify the improvement achieved when adapting a pretrained PEF model to new, unseen vasculatures, measuring the relationship between additional training time and performance gain. Two agents were fine-tuned from the PEF + adaptive horizon model: one for a right-sided ($A_1$, $A_{2R}$, $A_{3R}$) and one for a left-sided ($A_1$, $A_{2L}$, $A_{3L}$) navigation task.
                \item \textbf{\textit{In vitro} navigation of unseen vasculatures:} The PEF agent trained on the Train30--Test10 dataset was tested under fluoroscopy within an unseen \textit{in vitro} environment (selected from the hold-out set from the Train30--Test10 dataset). The effects of fine-tuning were also examined in these experiments. 
            \end{enumerate}
                
        \subsubsection{Evaluation}
    
            During \textit{in silico} training, evaluations were conducted every $2.5 \times 10^5$~exploration steps for $50$~episodes. Agents were allowed $450$~exploration steps (equivalent to approximately $60$\,s) during \textit{in silico} evaluation. The \textit{Success Rate} (SR) was defined as the percentage of evaluation episodes in which the controller successfully reaches the target; for unsuccessful episodes, \textit{Path Ratio} (PR) was defined as $100(d_0-d_T)/d_0$, where $d_0$ and $d_T$ are the initial and final centreline distances to the target, respectively; and \textit{Procedure Time} (PT) was defined as the time taken from the start of navigation to the target location for successful episodes~\cite{Robertshaw2026jaha}. Each evaluation step recorded the SR, PT, and PR. For the model with the highest SR, the corresponding PR and PT were examined. 
    
\begin{figure}[tb]
                \centering
                \includegraphics[width=0.55\linewidth]{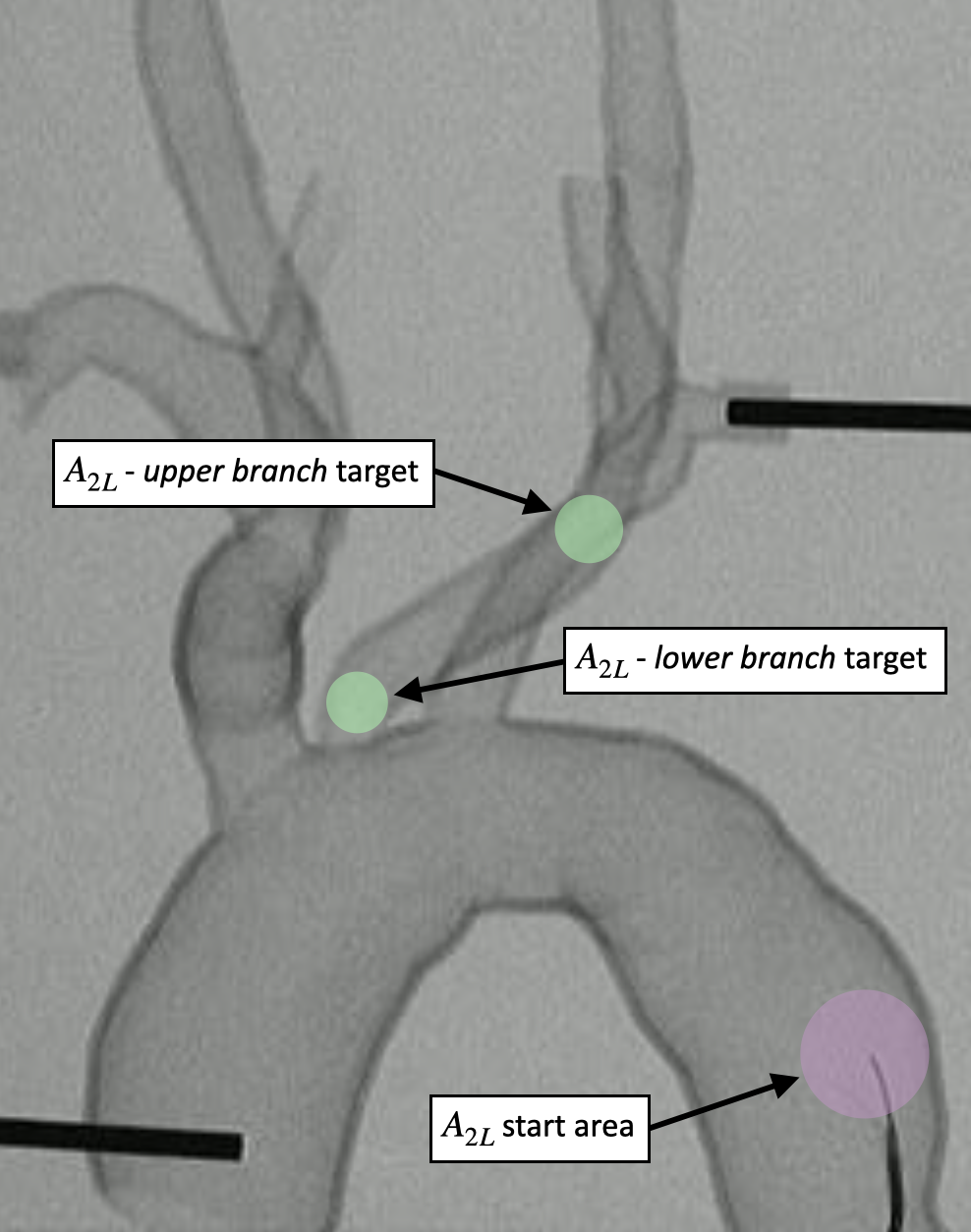}
                \caption{Phantom vessel model under fluoroscopy, showing task start area and target locations for $A_{2L}$ - \textit{lower branch} and $A_{2L}$ - \textit{upper branch}. Expert neurointerventionalists typically use an access catheter with an extreme shape (Simmons catheter) for this maneuver – this task can be considered a superhuman experiment~\cite{Robertshaw3_2025}.}
                \label{fig:a2ltasks}
            \end{figure}

            \textit{In vitro} testing was conducted in a mock operating room where each subtask was carried out five times for each agent, recording the SR, PT, and PR. For \textit{in vitro} evaluation of the Train30--Test10 dataset, the $A_{2L}$ subtask was split into two measures of success due to its difficulty~\cite{Robertshaw3_2025}, as shown in Fig.~\ref{fig:a2ltasks}. The first was navigating the guidewire inside of the LCCA branch to any degree ($A_{2L}$ - \textit{lower branch}), while the second was to navigate further up the LCCA to the original target ($A_{2L}$ - \textit{upper branch}, whereby the guidewire provides a stable base for the catheter to follow). Agents were allowed $5$\,\unit{\minute} during \textit{in vitro} evaluation, to account for the speed of the robotic manipulator.

            Comparative statistical analyses were conducted using two-sided independent-samples Student's $t$-tests, with statistical significance defined as $p<0.05$, for both \textit{in silico} and \textit{in vitro} experiments.

\section{RESULTS}

    \subsection{Experiment 1: \textit{In silico} validation on the Train10--Test0 dataset using PEF}

        Table~\ref{tab:insilico10.0} presents the \textit{in silico} performance of SAC, base TD-MPC2, and the proposed PEF framework across all navigation tasks using a dataset of 10 known real patient vasculatures. Overall, PEF achieved a mean SR of 74\% with improved performance compared to SAC ($37\%$, $p<0.001$), and a trend for improved performance compared to base TD-MPC2 ($65\%$, $p=0.053$). An improvement was seen when moving from base TD-MPC2 to PEF for $A_{3L}$ ($50\%$ to $94\%$, $p<0.001$). 
        
        PEF's increased SR appeared to be at the cost of an increase in mean PT ($23.8$\,\si{\second}) when compared to both model types (SAC: $8.2$\,\si{\second}, $p<0.001$; base TD-MPC2: $15.2$\,\si{\second}, $p<0.001$). The absolute time difference, however, is trivial compared to the duration of an MT ($\approx 64$\,\unit{\minute}~\cite{Weddell2020}). PEF achieved the highest mean PR ($87\%$), while also showing increases when compared to base TD-MPC2 across individual tasks $A_{2L}$ ($45\%$ to $75\%$, $p<0.001$) and $A_{3L}$ ($66\%$ to $92\%$, $p<0.001$).
    
            \begin{table*}[tb]
                \centering
                \scriptsize
                \caption{\textit{In silico} results per navigation task for a dataset of 10 real patient vasculatures, with mean results across all tasks and vasculatures. SAC and base TD-MPC2 results taken from \cite{Robertshaw3_2025}. Bold indicates the best value for each task (highest SR and PR; lowest PT).}
                \label{tab:insilico10.0}
                \begin{tabular}{c||ccc||ccc||ccc}
                    \multirow{2}{*}{Task} & \multicolumn{3}{c||}{SAC} & \multicolumn{3}{c||}{Base TD-MPC2} & \multicolumn{3}{c}{PEF} \\ \cline{2-10} 
                     & \multicolumn{1}{c|}{SR (\%)} & \multicolumn{1}{c|}{PT (s)} & PR (\%) & \multicolumn{1}{c|}{SR (\%)} & \multicolumn{1}{c|}{PT (s)} & PR (\%) & \multicolumn{1}{c|}{SR (\%)} & \multicolumn{1}{c|}{PT (s)} & PR (\%) \\ \hline
                    $A_1$ & \multicolumn{1}{l|}{96 $\pm$ 20} & \multicolumn{1}{l|}{\textbf{8.9 $\pm$ 1.6}} & 89 $\pm$ 16 & \multicolumn{1}{l|}{96 $\pm$ 20} & \multicolumn{1}{l|}{16.7 $\pm$ 4.1} & 90 $\pm$ 3 & \multicolumn{1}{l|}{\textbf{100 $\pm$ 0}} & \multicolumn{1}{l|}{12.3 $\pm$ 2.1} & $-$ \\ \hline
                    $A_{2L}$ & \multicolumn{1}{l|}{$10 \pm 30$} & \multicolumn{1}{l|}{\textbf{14.5 $\pm$ 7.1}} & $27 \pm 30$ & \multicolumn{1}{l|}{\textbf{22 $\pm$ 41}} & \multicolumn{1}{l|}{$17.0 \pm 6.1$} & 45 $\pm$ 33 & \multicolumn{1}{l|}{8 $\pm$ 27} & \multicolumn{1}{l|}{14.7 $\pm$ 6.0} & \textbf{75 $\pm$ 8} \\ \hline
                    $A_{2R}$ & \multicolumn{1}{l|}{$8 \pm 27$} & \multicolumn{1}{l|}{\textbf{12.8 $\pm$ 9.9}} & $29 \pm 27$ & \multicolumn{1}{l|}{66 $\pm$ 47} & \multicolumn{1}{l|}{$16.5 \pm 6.2$} & 77 $\pm$ 32 & \multicolumn{1}{l|}{\textbf{72 $\pm$ 45}} & \multicolumn{1}{l|}{28.6 $\pm$ 15.8} & \textbf{79 $\pm$ 29} \\ \hline
                    $A_{3L}$ & \multicolumn{1}{l|}{$16 \pm 37$} & \multicolumn{1}{l|}{\textbf{7.2 $\pm$ 7.0}} & $26 \pm 34$ & \multicolumn{1}{l|}{50 $\pm$ 50} & \multicolumn{1}{l|}{$13.7 \pm 6.3$} & 66 $\pm$ 38 & \multicolumn{1}{l|}{\textbf{94 $\pm$ 24}} & \multicolumn{1}{l|}{13.6 $\pm$ 8.6} & \textbf{92 $\pm$ 18} \\ \hline
                    $A_{3R}$ & \multicolumn{1}{l|}{$56 \pm 50$} & \multicolumn{1}{l|}{\textbf{5.6 $\pm$ 4.5}} & $63 \pm 43$ & \multicolumn{1}{l|}{90 $\pm$ 30} & \multicolumn{1}{l|}{$12.9 \pm 5.8$} & 90 $\pm$ 21 & \multicolumn{1}{l|}{\textbf{98 $\pm$ 14}} & \multicolumn{1}{l|}{42.7 $\pm$ 10.4} & \textbf{94 $\pm$ 2} \\ \hline\hline
                    Mean & \multicolumn{1}{l|}{37 $\pm$ 48} & \multicolumn{1}{l|}{\textbf{8.2 $\pm$ 4.7}} & \multicolumn{1}{l||}{47 $\pm$ 40} & \multicolumn{1}{l|}{65 $\pm$ 48} & \multicolumn{1}{l|}{15.2 $\pm$ 5.9} & \multicolumn{1}{l||}{73 $\pm$ 33} & \multicolumn{1}{l|}{\textbf{74 $\pm$ 44}} & \multicolumn{1}{l|}{23.8 $\pm$ 16.1} & \multicolumn{1}{l}{\textbf{87 $\pm$ 17}} \\
                \end{tabular}
            \end{table*}

    \subsection{Experiment 2: \textit{In vitro} validation on a Train10--Test0 vasculature using PEF}

        To evaluate \textit{in silico} to \textit{in vitro} domain transfer, models trained on the Train10–Test0 dataset were validated on a single known vasculature (\textit{in vitro}), as shown in Table~\ref{tab:invitro10.0}. PEF achieved a mean SR of 73\% with a trend for improved performance compared to SAC ($50\%$, $p=0.069$); and a similar performance compared to base TD-MPC2 ($63\%$, $p=0.414$), with comparable PTs to base TD-MPC2. Numerically, PEF recorded the equal or the highest SR across all tasks. PEF also achieved the highest mean PR ($73\%$) compared to SAC ($51\%$, $p<0.001$) and base TD-MPC2 ($61\%$, $p<0.001$). Importantly, both TD-MPC2-trained models (base TD-MPC2 and PEF) were frequently able to complete the $A2_L$ - \textit{lower branch} task. Here, PEF showed a superior SR of $60$\% compared to $0\%$ for SAC ($p=0.041$), and a similar SR compared to $40$\% for base TD-MPC2 ($p=0.581$).

        \begin{table*}[tb]
            \centering
            \scriptsize
            \caption{\textit{In vitro} results per navigation task for training on a dataset of 10 real patient vasculatures and testing on a single known vasculature, with mean results across tasks. SAC and base TD-MPC2 results taken from previous work, but updated for the splitting of the $A_{2L}$ task~\cite{robertshaw2026iros}. Bold indicates the best value for each task (highest SR and PR; lowest PT), `$-$' denotes cell is not applicable (PT is not recorded when SR is 0\%, and PR is not recorded when SR is 100\%).}
            \label{tab:invitro10.0}
            \begin{tabular}{c||ccc||ccc||ccc}
                \multirow{2}{*}{Task} & \multicolumn{3}{c||}{SAC} & \multicolumn{3}{c||}{Base TD-MPC2} & \multicolumn{3}{c}{PEF} \\ \cline{2-10} 
                & \multicolumn{1}{c|}{SR (\%)} & \multicolumn{1}{c|}{PT (s)} & \multicolumn{1}{c||}{PR (\%)} & \multicolumn{1}{c|}{SR (\%)} & \multicolumn{1}{c|}{PT (s)} & \multicolumn{1}{c||}{PR (\%)} & \multicolumn{1}{c|}{SR (\%)} & \multicolumn{1}{c|}{PT (s)} & \multicolumn{1}{c}{PR (\%)} \\ \hline
                $A_1$ & \multicolumn{1}{l|}{\textbf{100 $\pm$ 0}} & \multicolumn{1}{l|}{\textbf{46.2 $\pm$ 3.6}} & $-$ & \multicolumn{1}{l|}{\textbf{100 $\pm$ 0}} & \multicolumn{1}{l|}{$80.4 \pm 19.7$} & $-$ & \multicolumn{1}{l|}{\textbf{100 $\pm$ 0}} & \multicolumn{1}{l|}{80.6 $\pm$ 10.1} & $-$ \\ \hline
                $A_{2L}$ \textit{(lower)} & \multicolumn{1}{l|}{0 $\pm$ 0} & \multicolumn{1}{c|}{$-$} & $70 \pm 20$ & \multicolumn{1}{l|}{40 $\pm$ 55} & \multicolumn{1}{c|}{175.5 $\pm $0.7} & 82 $\pm$ 7 & \multicolumn{1}{l|}{\textbf{60 $\pm$ 55}} & \multicolumn{1}{c|}{\textbf{173.7 $\pm$ 81.1}} & \textbf{82 $\pm$ 6} \\ \hline
                $A_{2L}$ \textit{(upper)} & \multicolumn{1}{l|}{\textbf{0 $\pm$ 0}} & \multicolumn{1}{c|}{$-$} & $47 \pm 11$ & \multicolumn{1}{l|}{\textbf{0 $\pm$ 0}} & \multicolumn{1}{c|}{$-$} & 71 $\pm$ 14 & \multicolumn{1}{l|}{\textbf{0 $\pm$ 0}} & \multicolumn{1}{c|}{$-$} & \textbf{72 $\pm$ 5} \\ \hline
                $A_{2R}$ & \multicolumn{1}{l|}{$40 \pm 55$} & \multicolumn{1}{l|}{\textbf{97.0 $\pm$ 1.4}} & $42 \pm 4$ & \multicolumn{1}{l|}{60 $\pm$ 55} & \multicolumn{1}{l|}{$125.0 \pm 21.0$} & 49 $\pm$ 7 & \multicolumn{1}{l|}{\textbf{80 $\pm$ 45}} & \multicolumn{1}{l|}{127 $\pm$ 51} & \textbf{66 $\pm$ 0} \\ \hline
                $A_{3L}$ & \multicolumn{1}{l|}{\textbf{100 $\pm$ 0}} & \multicolumn{1}{l|}{\textbf{54.4 $\pm$ 0}} & $-$ & \multicolumn{1}{l|}{\textbf{100 $\pm$ 0}} & \multicolumn{1}{l|}{$126.4 \pm 57.7$} & $-$ & \multicolumn{1}{l|}{\textbf{100 $\pm$ 0}} & \multicolumn{1}{l|}{$109.0 \pm 42.4$} & $-$ \\ \hline
                $A_{3R}$ & \multicolumn{1}{l|}{\textbf{60 $\pm$ 55}} & \multicolumn{1}{l|}{\textbf{67.0 $\pm$ 7.2}} & 45 $\pm$ 30 & \multicolumn{1}{l|}{80 $\pm$ 45} & \multicolumn{1}{l|}{$112.3 \pm 7.5$} & $43 \pm 0$ & \multicolumn{1}{l|}{\textbf{100 $\pm$ 0}} & \multicolumn{1}{l|}{$120.0 \pm 17.7$} & $-$\\ \hline\hline
                Mean & \multicolumn{1}{l|}{$50 \pm 51$} & \multicolumn{1}{l|}{\textbf{66.2 $\pm$ 17.2}} & $51 \pm 19$ & \multicolumn{1}{l|}{63 $\pm$ 49} & \multicolumn{1}{l|}{$123.9 \pm 40.9$} & 61 $\pm$ 17 & \multicolumn{1}{l|}{\textbf{73 $\pm$ 45}} & \multicolumn{1}{l|}{122.1 $\pm$ 47.3} & \textbf{73 $\pm$ 7} \\
            \end{tabular}
        \end{table*}

    \subsection{Experiment 3: \textit{In silico} generalization on the Train30--Test10 dataset using PEF with an adaptive horizon}

        Evaluation results from a PEF model trained on 30 real patient vasculatures and evaluated on 10 unseen anatomies are shown in Table~\ref{tab:insilico30.10}. The baseline PEF achieved a mean SR of $89\%$, a mean PR of $69\%$, and a mean PT of $20.6$\,\unit{\second}. Utilizing our \textit{adaptive horizon} method with PEF gave a similar mean SR of $90\%$ ($p=0.719$) with an improved PR of $78\%$ ($p<0.001$), and a similar mean PT of $19.5$\,\si{\second} ($p=0.416$). Similar SR was observed across $A_{1}$, $A_{2R}$, $A_{3L}$, and $A_{3R}$. On the most difficult task ($A_{2L}$; Fig.~\ref{fig:a2ltasks}), the SR with an adaptive horizon was $60\%$ compared to $52\%$ ($p=0.421$) with PR increasing from $69\%$ to $78\%$ ($p=0.031$).

        \begin{table}[tb]
            \centering
            \scriptsize
            \setlength{\tabcolsep}{2.5pt} 
            \caption{\textit{In silico} results per navigation task for 30 training and 10 unseen patient vasculatures. Bold indicates the best value for each task (highest SR and PR; lowest PT), `$-$' denotes cell is not applicable (PT is not recorded when SR is 0\%, and PR is not recorded when SR is 100\%).}
            \label{tab:insilico30.10}
            \begin{tabular}{c||c|c|c||c|c|c}
                \multirow{2}{*}{Task} 
                & \multicolumn{3}{c||}{PEF} 
                & \multicolumn{3}{c}{PEF + Adaptive Horizon} \\ \cline{2-7}
                & SR (\%) & PT (s) & PR (\%) & SR (\%) & PT (s) & PR (\%) \\ \hline
                $A_1$   & \textbf{100$\pm$0} & \textbf{8.0$\pm$0.6} & $-$ 
                        & \textbf{100$\pm$0} & $10.0\pm1.8$ & $-$ \\ \hline
                $A_{2L}$ & $52\pm50$ & $35.6\pm16.0$ & $69\pm14$
                         & \textbf{60$\pm$49} & \textbf{28.5$\pm$17.5} & \textbf{79$\pm$13} \\ \hline
                $A_{2R}$ & \textbf{96$\pm$20} & $26.5\pm13.4$ & \textbf{77$\pm$26}
                         & $92\pm27$ & \textbf{24.6$\pm$14.7} & $72\pm3$ \\ \hline
                $A_{3L}$ & \textbf{100$\pm$0} & \textbf{19.5$\pm$14.2} & $-$
                         & $98\pm14$ & $20.2\pm14.0$ & \textbf{76$\pm$0} \\ \hline
                $A_{3R}$ & $96\pm20$ & $20.7\pm16.6$ & $62\pm19$
                         & \textbf{100$\pm$0} & \textbf{18.4$\pm$15.8} & $-$ \\ \hline\hline
                Mean & $89\pm32$ & $20.6\pm15.5$ & $69\pm15$
                     & \textbf{90$\pm$30} & \textbf{19.5$\pm$14.7} & \textbf{78$\pm$12} \\
            \end{tabular}
        \end{table}

    \subsection{Experiment 4: Fine-tuning on unseen \textit{in silico} vasculature using PEF with an adaptive horizon}\label{ablation}

\begin{figure}[tb]
            \centering
            \includegraphics[width=0.9\linewidth]{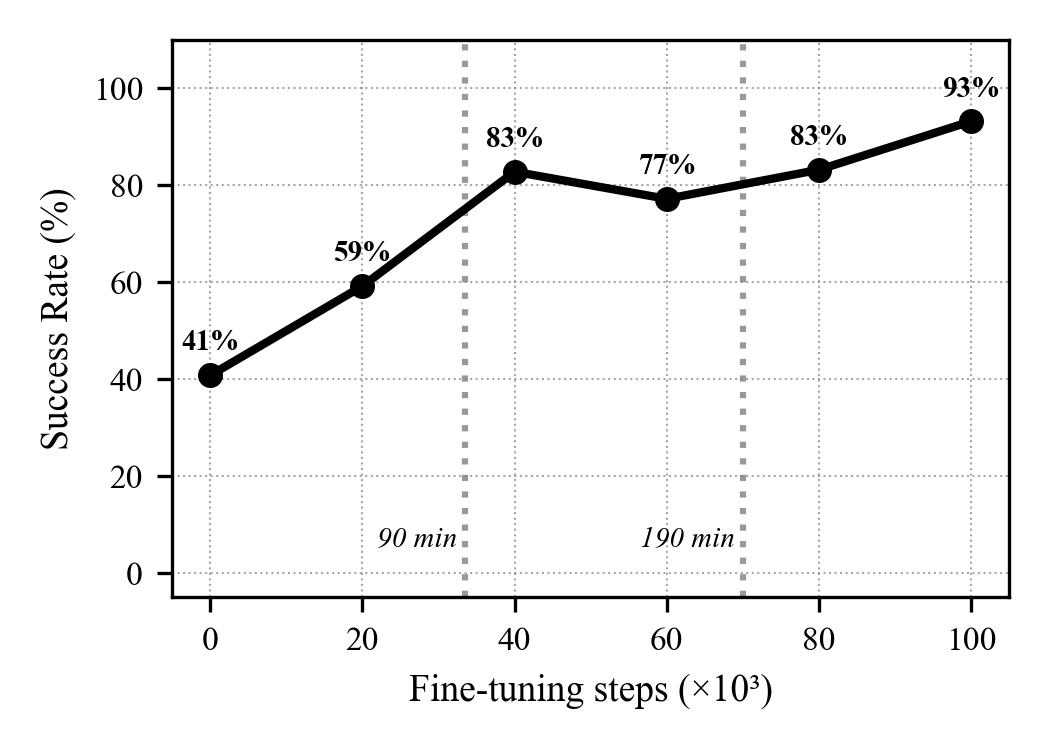}
            \caption{Plot of mean task success rate vs number of fine-tuning steps for a single \textit{in silico} vasculature using PEF and an adaptive horizon. Training times in relation to key times are highlighted: $90$\,\unit{\minute} = $34\times 10^3$~steps, $190$\,\unit{\minute} = $70\times 10^3$~steps}
            \label{fig:ft_graph}
        \end{figure}

\begin{figure}[tb]
            \centering
            \includegraphics[width=0.9\linewidth]{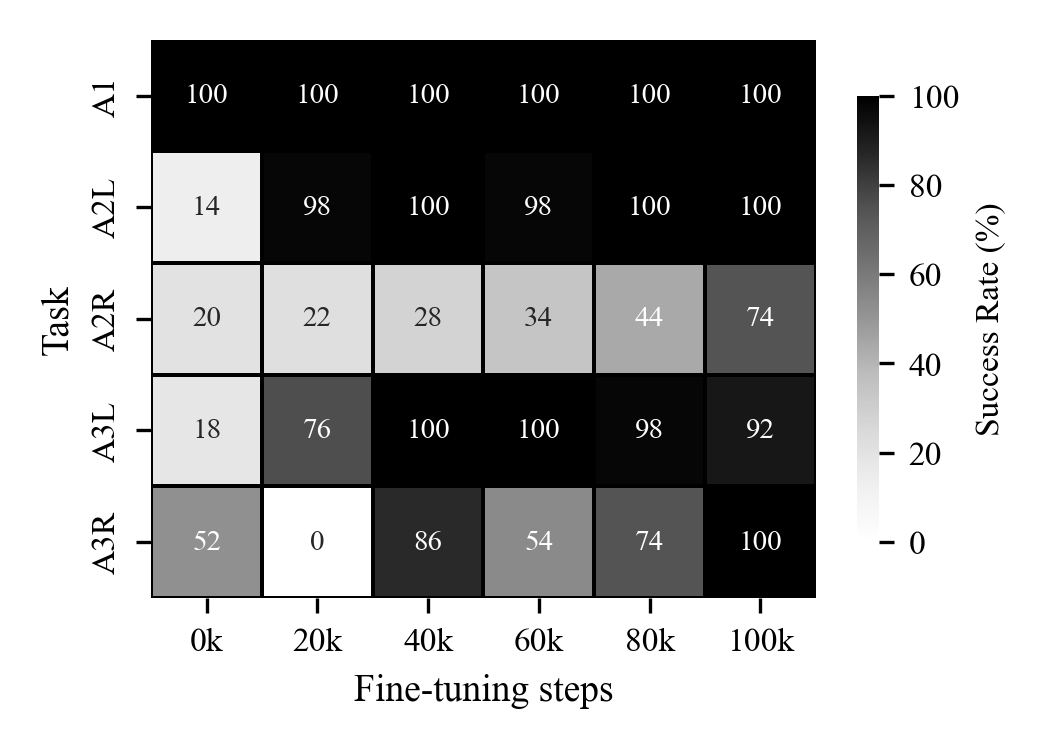}
            \caption{Heatmap of individual task success rate vs number of fine-tuning steps for a single \textit{in silico} vasculature using PEF and an adaptive horizon. $20\times 10^3$ fine-tuning steps $\approx 54$\,\unit{\minute}.}
            \label{fig:ft_heatmap}
        \end{figure}

        During \textit{in silico} fine-tuning, training speeds were calculated to be $\approx 6.21$\,\unit{{steps}\per\second}, meaning $20\times 10^3$ fine-tuning steps were equivalent to $\approx 54$\,\unit{\minute}. Inter-hospital transfer times have been shown to be $\approx 190$\,\unit{\minute}, and for patients presenting initially at MT-capable centers it takes $90$\,\unit{\minute} from CT scan to groin puncture~\cite{Sun2013}, meaning on average, fine-tuning could take place for $34$ to $70\times 10^3$ steps. Fig.~\ref{fig:ft_graph} shows the effect on task performance that increased fine-tuning steps had on mean task SR, while Fig.~\ref{fig:ft_heatmap} shows a heatmap of individual task performance. $40\times 10^3$ fine-tuning steps increased the mean SR across tasks from $41\%$ to $83\%$ ($p<0.001$), while $100\times 10^3$ fine-tuning steps further increased the mean SR to $93\%$, an improvement over the $40\times 10^3$ model ($p<0.001$). 

    \subsection{Experiment 5: \textit{In vitro} navigation of unseen vasculature}

        Table~\ref{tab:invitro30.10} presents navigation results using 30 vasculatures tested on an unseen \textit{in vitro} vasculature (as with all experiments these represent real diverse patient vasculatures). Based on the results of Section~\ref{ablation}, the $40\times 10^3$ fine-tuning step model was used for comparison during evaluation. Fine-tuning gave a mean SR of $67\%$ compared to $60\%$ without fine-tuning ($p=0.623$). Mean PR improved from $63\%$ to $80\%$ with fine-tuning ($p<0.001$), while mean PT did not appear to be compromised. Both models were able to frequently carry out the $A_{2L}$ - \textit{lower branch} task. Here, the addition of fine-tuning gave a mean SR of $40\%$ compared to $20\%$ ($p=0.547$).

        \begin{table}[tb]
            \centering
            \scriptsize
            \setlength{\tabcolsep}{2.5pt} 
            \caption{\textit{In vitro} results per navigation task for 30 training and one unseen patient vasculature. Bold indicates the best value for each task (highest SR and PR; lowest PT), `$-$' denotes cell is not applicable (PT is not recorded when SR is 0\%, and PR is not recorded when SR is 100\%).}
            \label{tab:invitro30.10}
            \begin{tabular}{c||c|c|c||c|c|c}
                \multirow{2}{*}{Task} 
                & \multicolumn{3}{c||}{\begin{tabular}[c]{@{}c@{}} PEF \\ + Adaptive horizon \end{tabular}} 
                & \multicolumn{3}{c}{\begin{tabular}[c]{@{}c@{}} PEF \\ + Adaptive horizon \\ + Fine-tuning \end{tabular}} \\ \cline{2-7}
                & SR (\%) & PT (s) & PR (\%) & SR (\%) & PT (s) & PR (\%) \\ \hline
                $A_1$ & \textbf{100$\pm$0} & \textbf{53.8$\pm$4.8} & $-$ 
                     & \textbf{100$\pm$0} & $61.0\pm18.2$ & $-$ \\ \hline
                \begin{tabular}[c]{@{}c@{}} $A_{2L}$ \\(\textit{lower}) \end{tabular}  & $20\pm45$ & $157\pm0$ & $77\pm12$
                     & \textbf{40$\pm$55} & \textbf{136.5$\pm$40.3} & \textbf{91$\pm$2} \\ \hline
                \begin{tabular}[c]{@{}c@{}} $A_{2L}$ \\(\textit{upper}) \end{tabular} & $0\pm0$ & $-$ & $67\pm10$
                     & $0\pm0$ & $-$ & \textbf{75$\pm$3} \\ \hline
                $A_{2R}$ & \textbf{60$\pm$55} & \textbf{167.7$\pm$64.4} & $73\pm3$
                     & \textbf{60$\pm$55} & $175.7\pm92.6$ & \textbf{74$\pm$4} \\ \hline
                $A_{3L}$ & $80\pm45$ & $81.3\pm20.0$ & $37\pm0$
                     & \textbf{100$\pm$0} & \textbf{57.0$\pm$9.2} & $-$ \\ \hline
                $A_{3R}$ & \textbf{100$\pm$0} & \textbf{103.6$\pm$50.0} & $-$
                     & \textbf{100$\pm$0} & $137.6\pm25.3$ & $-$ \\ \hline\hline
                Mean & $60\pm50$ & \textbf{112.7$\pm$53.5} & $63\pm14$
                     & \textbf{67$\pm$50} & $113.6\pm59.1$ & \textbf{80$\pm$8} \\
            \end{tabular}
        \end{table}

\begin{figure}[tb]
            \centering
            \includegraphics[width=0.85\linewidth]{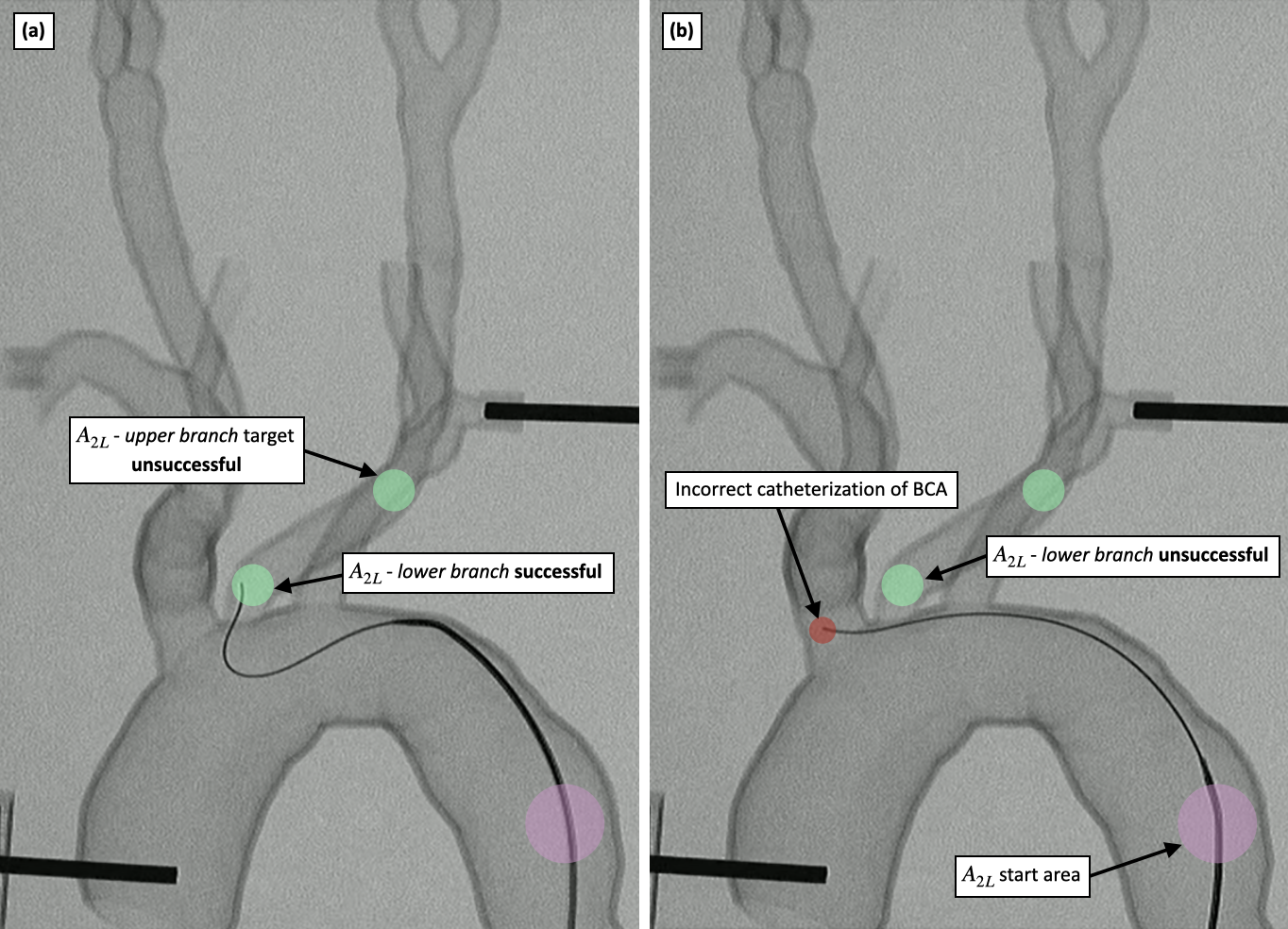}
            \caption{Vessel model under fluoroscopy, showing \textbf{(a)} successful navigation of $A_{2L}$ - \textit{lower branch}, and \textbf{(b)} unsuccessful navigation of $A_{2L}$ - \textit{lower branch} due to catheterization of the BCA (Brachiocephalic Artery).}
            \label{fig:a2l_diagram}
        \end{figure}

\begin{figure}[tb]
            \centering
            \includegraphics[width=0.85\linewidth]{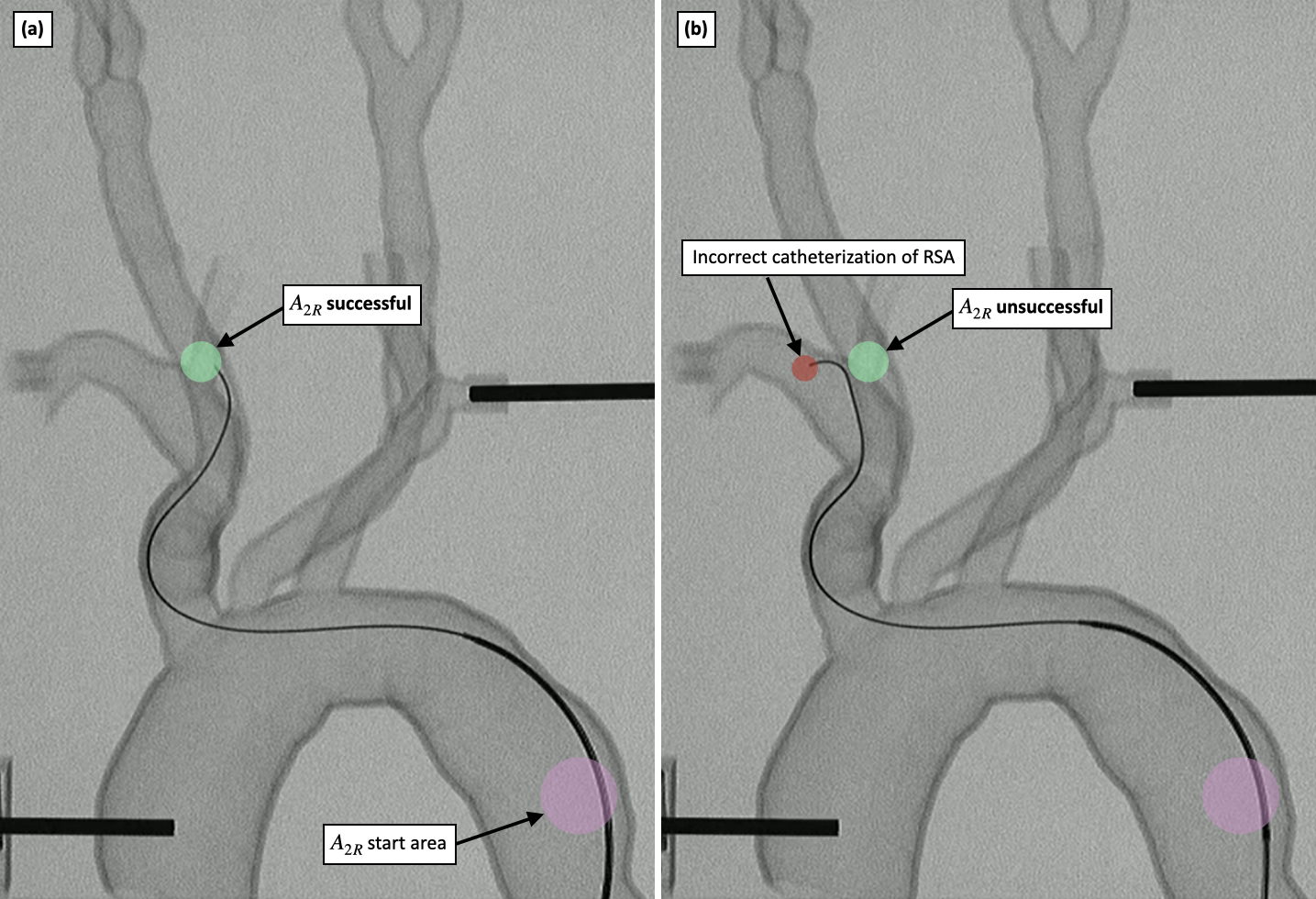}
            \caption{Vessel model under fluoroscopy, showing \textbf{(a)} successful navigation of $A_{2R}$, and \textbf{(b)} unsuccessful navigation of $A_{2R}$ due to catheterization of the RSA (Right Subclavian Artery).}
            \label{fig:a2r_diagram}
        \end{figure}

        Fig.~\ref{fig:a2l_diagram} shows annotated fluoroscopic images of a successful \textit{in vitro} navigation attempt of $A_{2L}$ - \textit{lower branch}, and an unsuccessful navigation of $A_{2L}$ - \textit{lower branch} due to catheterization of the brachiocephalic artery. Additionally, Fig.~\ref{fig:a2r_diagram} shows fluoroscopic images of a successful \textit{in vitro} navigation attempt of $A_{2R}$, and an unsuccessful navigation of $A_{2R}$ due to catheterization of the right subclavian artery.

\section{DISCUSSION}

    To our knowledge, this study provides the first completion of autonomous endovascular navigation under fluoroscopy in a previously unseen, patient-derived \textit{in vitro} stroke vasculature. The proposed PEF framework enabled a single world-model controller to perform multiple clinically-relevant navigation tasks across different anatomies, with successful transfer from \textit{in silico} to \textit{in vitro}, while showing improvements over previous state-of-the-art methods. The demonstration of the first autonomous navigation of an unseen \textit{in vitro} stroke patient vasculature is a key step towards the benefits of fully autonomous MT.

    \subsection{Progressive experience fusion}

        PEF provided consistent improvements in SR and PR compared with previously proposed SAC and base TD-MPC2 configurations, supporting the proposal that the fusing of replay buffers from multiple single-task agents enables scalable multi-task learning, without requiring computationally expensive multi-task and multi-anatomy training from scratch. While PEF successfully scales to five varying navigation tasks in this study, further work is needed to evaluate its scalability to full MT navigation and additional clinical variations (e.g., highly tortuous vessels). Increasing the number of parameters of the world model may provide benefits for these extended tasks, as previous TD-MPC2 research has shown that larger models (e.g., 19M parameters) exhibit improved outcomes, although at higher computational cost~\cite{Hansen2024}. This trade-off between performance and training complexity must be balanced for future clinical translation; however, it is hoped that the proposed PEF framework will reduce the training burden.

        While this study focused on the first stage of MT, the full intervention is multi-device and multi-stage. In our current implementation of PEF, we consider a $D=2$ task space defined by task ($N_{\mathrm{task}}$) and environment ($N_{\mathrm{env}}$), yielding $M_0=N_{\mathrm{env}}N_{\mathrm{task}}$ single-task agents. However, PEF is scalable to higher-dimensional structures. Additional dimensions could represent device type, patient data modality, or procedural phase. Each additional dimension would introduce another fusion layer, although scalability beyond the two dimensions evaluated here remains to be established. While this work focuses on navigation, the same hierarchical structure could accommodate downstream endovascular intervention tasks such as clot retrieval or stent deployment as additional task layers. The concept is plausible given that navigation in the neck vasculature is arguably the most challenging aspect of MT and has been demonstrated here. The realization of such a generalizable agent would represent a significant advancement toward fully autonomous endovascular interventions.
        
        Furthermore, previous research in multi-task RL agents (not related to autonomous endovascular navigation) has typically been one-dimensional, whereby learned experiences of specialized agents are combined across tasks only~\cite{Hansen2024}. The proposed PEF instead fuses post-training experience sequentially across multiple sources of variation, which in this study were vascular anatomy and navigation subtask. We believe this multi-dimensional paradigm applies to a broad range of domains requiring hierarchical generalization. For example, other areas of surgical robotics (hierarchies can be formed over procedure type, anatomy, or surgical tool), and broader robotic manipulation (hierarchies can be formed over different objects, manipulation skills, or robot types).

    \subsection{Adaptive horizon}

        Our adaptive horizon implementation provided no improvements in mean SR \textit{in silico} ($89\%$ to $90\%$, $p = 0.719$), but did increase mean PR for the \textit{in silico} testbed ($69\%$ to $78\%$, $p < 0.001$), meaning that in failed navigations the addition of the adaptive horizon resulted in the agent finishing closer to the target. In this setting, failures are dominated by device-control constraints rather than planning horizon. Particularly, $A_{2L}$ - \textit{lower branch}, where the main limitation is the catheter failing to provide a stable base, a failure mode unaffected by longer planning. Tasks with planning-bound failures show the expected PR improvement (closer approach to target), but SR is gated by a binary device-kinematics barrier that adaptive horizon alone cannot cross. Future work should investigate other strategies for horizon adaptation to determine whether these PR improvements can translate to consistent SR gains. Additional evaluation on general RL benchmarks with varying task lengths is also required.        
        
    \subsection{Fine-tuning}

        Our proposed method of fine-tuning a single general agent demonstrated rapid adaptation to new unseen vasculature. Gains in performance became non-monotonic after approximately $40 \times 10^3$ fine-tuning steps, which equates to $107$\,\unit{\minute} and so is within a real-world inter-hospital transfer window, and is close to the CT-to-puncture window in MT-capable hospitals where patients are not transferred. All fine-tuning experiments were performed on a single NVIDIA A100 GPU, and the reported times reflect this computational setup. While such hardware may not be directly available in a clinical environment, fine-tuning primarily involves forward and backward passes through a compact latent world model, suggesting that future work could research if similar adaptation could be achieved on more modest GPU resources. 
        
\begin{figure}[tb]
            \centering
            \includegraphics[width=1\linewidth]{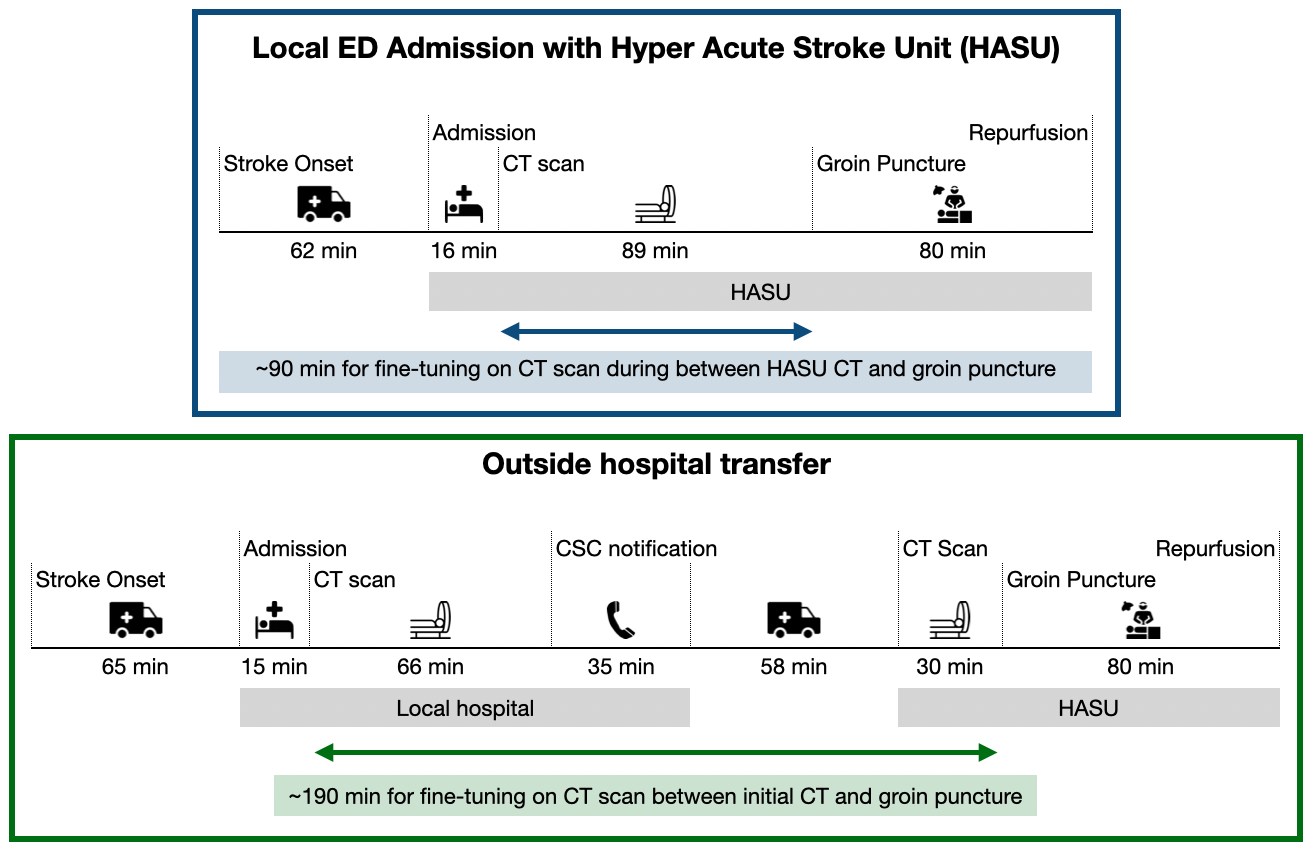}
            \caption{Patient timeline from stroke onset to reperfusion, showing potential benefit of patient-specific fine-tuning for local admission at an MT-capable center, and outside hospital transfer to an MT-capable center. Data from \cite{Sun2013}.}
            \label{fig:timeline}
        \end{figure}

        These results align with non-medical application-based studies that have shown pretrained TD-MPC2 world models can be adapted to unseen tasks with few-shot interaction~\cite{Hansen2024}. In MT, this capability directly maps to clinical workflows where patient imaging becomes available before treatment through teleradiology, potentially enabling personalized autonomous navigation strategies before reaching the MT-capable center. Time from initial CT scan to groin puncture has been shown to be $\approx 90$\,\unit{\minute} at MT-capable centers, and $\approx 190$\,\unit{\minute} at sites where MT is not available~\cite{Sun2013}. This provides a clinically relevant time window for the agent to fine-tune its latent dynamics representation and action policy before beginning the intervention, as shown in Fig.~\ref{fig:timeline}. 

        Patient-specific simulation may also support intervention-time estimation and identify difficult navigation segments before treatment. Failure to complete the \textit{in silico} route could provide a pre-procedural indication that full control should be retained by the expert interventional neuroradiologist.
    
    \subsection{Subtasks}

        The $A_{2L}$ navigation task (navigation from the top of the descending aorta to a stable point with the LCCA) remains the most challenging subtask due to complex branching geometry and difficult vessel take-off angles from the aortic arch. Splitting the task into the lower and upper LCCA branches for evaluation purposes revealed that partial entry was achieved, but the inability to effectively utilize the catheter to provide a stable base halted navigation progress further up the LCCA. It is possible that by using dual-tracking methods, these issues could be mitigated~\cite{Robertshaw2024}, but effective live tip tracking of both guidewire and catheter \textit{in vitro} remains difficult. An alternative method to enable effective catheter usage could be through image-based RL, whereby the entire fluoroscopic image is used as an input at every step, and hence the agent would be able to see the catheter in the image. However, utilizing an image with high enough resolution to capture the visual differences between the guidewire and catheter would add significant computational burden and complexity to RL training. Our model-based approach learns latent dynamics, meaning it may be possible to implicitly capture catheter–guidewire interactions and compensate for partial observability without requiring high-dimensional image inputs. Stability during LCCA navigation could be improved by extending the model to better represent these interactions.

    \subsection{Limitations}
    
        Only two types of device were used throughout this study, and it is hoped that by including more devices SRs on navigation tasks in complex vasculatures can be increased. While we aimed to perform a task that can be considered a superhuman experiment in the current work, future work using enabling devices employed by experts would plausibly achieve the challenging $A_{2L}$ task, for example. Additionally, only a single unseen \textit{in vitro} phantom was studied in this paper; hence, future work should aim to evaluate a wider range of unseen \textit{in vitro} anatomies, before moving towards \textit{ex vivo} and \textit{in vivo} modalities. While the \textit{in vitro} phantom was able to capture the complex navigation path of stroke patient vasculature, additional complexity could be added by utilizing less rigid vessels and simulated pulsatility~\cite{Robertshaw2026jaha}. The present results establish proof of concept for fluoroscopy-guided transfer to one held-out patient-derived geometry. However, demonstrating that unseen \textit{in vitro} vascular navigation can be completed under fluoroscopy marks an important milestone in reducing geographic and expertise-related inequities in stroke care delivery.

\section{CONCLUSION}

    This work demonstrated, for the first time, autonomous endovascular navigation in a previously unseen \textit{in vitro} stroke patient vasculature using a world model-based RL approach. Using the proposed PEF framework, a single TD-MPC2 agent was trained to perform multiple navigation tasks across diverse vasculatures, achieving higher performance when compared to state-of-the-art baselines. The addition of adaptive horizon planning to TD-MPC2 decreased distance to target in failed attempts, while patient-specific fine-tuning enabled continued training through simulated interactions within clinically realistic inter-hospital transfer times. Together, these results represent a significant step toward scalable autonomous robotic assistance for MT, and provide a foundation for extending autonomy to the full procedural workflow in future work.

\section*{Acknowledgments}

    The authors would like to thank the staff and facilities at Surgical \& Interventional Engineering, King's College London, for their support, particularly Duane James and Gayathri Nantharatnam for their assistance during the experimental work.

\bibliographystyle{IEEEtran} 
\bibliography{IEEEexample}

@article{Robertshaw2023,
   author = {Harry Robertshaw and others},
   doi = {10.3389/fnhum.2023.1239374},
   issn = {1662-5161},
   journal = {Frontiers in Human Neuroscience},
   month = {8},
   title = {Artificial intelligence in the autonomous navigation of endovascular interventions: a systematic review},
   volume = {17},
   year = {2023},
}

@article{Pu2023,
   author = {Liyuan Pu and Li Wang and Ruijie Zhang and Tian Zhao and Yannan Jiang and Liyuan Han},
   doi = {10.1161/STROKEAHA.122.040073},
   issn = {15244628},
   issue = {5},
   journal = {Stroke},
   month = {5},
   pages = {1330-1339},
   pmid = {37094034},
   publisher = {Wolters Kluwer Health},
   title = {Projected Global Trends in Ischemic Stroke Incidence, Deaths and Disability-Adjusted Life Years From 2020 to 2030},
   volume = {54},
   year = {2023}
}

@article{Feigin2021,
   author = {Valery L Feigin and others},
   doi = {10.1016/S1474-4422(21)00252-0},
   issn = {14744422},
   journal = {The Lancet Neurology},
   title = {Global, regional, and national burden of stroke and its risk factors, 1990–2019: a systematic analysis for the Global Burden of Disease Study 2019},
   volume = {20},
   year = {2021}
}

@misc{Kazi2024,
   author = {Dhruv S. Kazi and others},
   doi = {10.1161/CIR.0000000000001258},
   issn = {15244539},
   issue = {4},
   journal = {Circulation},
   month = {7},
   pages = {e89-e101},
   pmid = {38832515},
   publisher = {Lippincott Williams and Wilkins},
   title = {Forecasting the Economic Burden of Cardiovascular Disease and Stroke in the United States Through 2050: A Presidential Advisory From the American Heart Association},
   volume = {150},
   year = {2024}
}

@article{Zhang2025,
   author = {Renfei Zhang and Linjie Dong and Xingsong Wang and Mengqian Tian and Haobo Su},
   doi = {10.1109/TIM.2025.3545167},
   issn = {15579662},
   journal = {IEEE Transactions on Instrumentation and Measurement},
   publisher = {Institute of Electrical and Electronics Engineers Inc.},
   title = {A Sensor-less Guider Contact Force Estimation Approach for Endovascular Slender Robot Assisted Guidance System},
   year = {2025}
}

@article{Yao2025,
   author = {Tianliang Yao and others},
   doi = {10.1109/TASE.2025.3555559},
   issn = {15583783},
   journal = {IEEE Transactions on Automation Science and Engineering},
   publisher = {Institute of Electrical and Electronics Engineers Inc.},
   title = {Sim2Real Learning with Domain Randomization for Autonomous Guidewire Navigation in Robotic-Assisted Endovascular Procedures},
   year = {2025}
}

@article{Bendszus2023,
   author = {Martin Bendszus and others},
   doi = {10.1016/S0140-6736(23)02032-9},
   issn = {1474547X},
   journal = {The Lancet},
   pmid = {37837989},
   publisher = {Elsevier B.V.},
   title = {Endovascular thrombectomy for acute ischaemic stroke with established large infarct: multicentre, open-label, randomised trial},
   year = {2023}
}

@article{Nogueira2018,
   author = {Raul G. Nogueira and others},
   doi = {10.1056/nejmoa1706442},
   issn = {0028-4793},
   issue = {1},
   journal = {New England Journal of Medicine},
   month = {1},
   pages = {11-21},
   pmid = {29129157},
   publisher = {Massachusetts Medical Society},
   title = {Thrombectomy 6 to 24 Hours after Stroke with a Mismatch between Deficit and Infarct},
   volume = {378},
   year = {2018}
}

@article{Asdaghi2023,
   author = {Negar Asdaghi and others},
   doi = {10.1161/STROKEAHA.122.040352},
   issn = {15244628},
   issue = {3},
   journal = {Stroke},
   month = {3},
   pages = {733-742},
   pmid = {36848428},
   publisher = {Wolters Kluwer Health},
   title = {Impact of Time to Treatment on Endovascular Thrombectomy Outcomes in the Early Versus Late Treatment Time Windows},
   volume = {54},
   year = {2023}
}

@article{Saver2016,
   author = {Jeffrey L. Saver and others},
   doi = {10.1001/jama.2016.13647},
   issn = {15383598},
   issue = {12},
   journal = {JAMA - Journal of the American Medical Association},
   month = {9},
   pages = {1279-1288},
   pmid = {27673305},
   publisher = {American Medical Association},
   title = {Time to treatment with endovascular thrombectomy and outcomes from ischemic stroke: A meta-analysis},
   volume = {316},
   year = {2016}
}

@techReport{SSNAP2023,
   author = {SSNAP},
   institution = {Sentinel Stroke National Audit Programme},
   title = {SSNAP Annual Report 2023},
   url = {www.hqip.org.uk/national-programmes},
   year = {2023}
}

@article{McMeekin2024,
   author = {Peter McMeekin and Martin James and Christopher I. Price and Gary A. Ford and Philip White},
   doi = {10.1177/23969873241232820},
   issn = {23969881},
   issue = {3},
   journal = {European Stroke Journal},
   month = {9},
   pages = {566-574},
   pmid = {38366958},
   publisher = {SAGE Publications Ltd},
   title = {The impact of large core and late treatment trials: An update on the modelled annual thrombectomy eligibility of UK stroke patients},
   volume = {9},
   year = {2024}
}

@article{Zhang2021,
   author = {Liqun Zhang and others},
   doi = {10.7861/CLINMED.2020-0579},
   issn = {14734893},
   issue = {1},
   journal = {Clinical Medicine, Journal of the Royal College of Physicians of London},
   month = {1},
   pages = {E26-E31},
   pmid = {33479080},
   publisher = {Royal College of Physicians},
   title = {Hub-and-spoke model for thrombectomy service in UK NHS practice},
   volume = {21},
   year = {2021}
}

@article{Regenhardt2018,
   author = {Robert W. Regenhardt and others},
   doi = {10.1161/STROKEAHA.118.020618},
   issn = {15244628},
   issue = {6},
   journal = {Stroke},
   pages = {1419-1425},
   pmid = {29712881},
   publisher = {Lippincott Williams and Wilkins},
   title = {Delays in the air or ground transfer of patients for endovascular thrombectomy},
   volume = {49},
   year = {2018}
}

@article{Klein2009,
   author = {Lloyd W Klein and others},
   issue = {2},
   journal = {Society of Interventional Radiology},
   month = {2},
   pages = {538-544},
   title = {Occupational Health Hazards in the Interventional Laboratory: Time for a Safer Environment},
   volume = {250},
   year = {2009}
}

@article{Madder2017,
   author = {Ryan D. Madder and others},
   doi = {10.1016/j.carrev.2016.12.011},
   issn = {18780938},
   issue = {3},
   journal = {Cardiovascular Revascularization Medicine},
   month = {4},
   pmid = {28041859},
   publisher = {Elsevier Inc.},
   title = {Impact of robotics and a suspended lead suit on physician radiation exposure during percutaneous coronary intervention},
   volume = {18},
   year = {2017}
}

@article{RiveroMoreno2024,
   author = {Yeisson Rivero-Moreno and others},
   doi = {10.7759/cureus.52243},
   journal = {Cureus},
   month = {1},
   publisher = {Springer Science and Business Media LLC},
   title = {Autonomous Robotic Surgery: Has the Future Arrived?},
   year = {2024}
}

@inproceedings{Yu2019,
   author = {Tianhe Yu and others},
   city = {Osaka, Japan},
   booktitle = {3rd Conference on Robot Learning (CoRL 2019)},
   title = {Meta-World: A Benchmark and Evaluation for Multi-Task and Meta Reinforcement Learning},
   year = {2019}
}

@article{Robertshaw2024,
   author = {Harry Robertshaw and Lennart Karstensen and Benjamin Jackson and Alejandro Granados and Thomas C. Booth},
   doi = {10.1007/s11548-024-03208-w},
   issn = {1861-6429},
   journal = {Int J CARS},
   month = {6},
   title = {Autonomous navigation of catheters and guidewires in mechanical thrombectomy using inverse reinforcement learning},
   year = {2024}
}

@article{Robertshaw1_2025,
   author = {Harry Robertshaw and others},
   doi = {10.1007/s11548-025-03339-8},
   issn = {18616429},
   journal = {Int J CARS},
   publisher = {Springer Science and Business Media Deutschland GmbH},
   title = {Reinforcement learning for safe autonomous two-device navigation of cerebral vessels in mechanical thrombectomy},
   year = {2025}
}

@inproceedings{Henderson2018,
   author = {Peter Henderson and Riashat Islam and Philip Bachman and Joelle Pineau and Doina Precup and David Meger},
   booktitle = {Thirty-Second AAAI Conference on Artificial Intelligence and Thirtieth Innovative Applications of Artificial Intelligence Conference and Eighth AAAI Symposium on Educational Advances in Artificial Intelligence},
   month = {2},
   title = {Deep Reinforcement Learning that Matters},
   year = {2018}
}

@article{Georgiev2024,
   author = {Ignat Georgiev and Varun Giridhar and Nicklas Hansen and Animesh Garg},
   month = {7},
   title = {PWM: Policy Learning with Large World Models},
   url = {http://arxiv.org/abs/2407.02466},
   year = {2024}
}

@article{Hafner2023,
   author = {Danijar Hafner and Jurgis Pasukonis and Jimmy Ba and Timothy Lillicrap},
   month = {1},
   title = {Mastering Diverse Domains through World Models},
   url = {http://arxiv.org/abs/2301.04104},
   year = {2023}
}

@article{Weddell2020,
   author = {Jake Weddell and others},
   doi = {10.1186/s12883-020-01909-8},
   issn = {14712377},
   issue = {1},
   journal = {BMC Neurology},
   month = {9},
   pmid = {32873250},
   publisher = {BioMed Central Ltd},
   title = {Mechanical thrombectomy: Can it be safely delivered out of hours in the UK?},
   volume = {20},
   year = {2020}
}

@misc{SSNAP2024,
   author = {{Sentinel Stroke National Audit Programme}},
   title = {SSNAP Annual Report 2024},
   url = {https://www.hqip.org.uk/resource/ssnap-nov24/},
   year = {2024},
   month = {11},
   day = {14},
   howpublished = {Healthcare Quality Improvement Partnership},
}

@article{Berkhemer2015,
   author = {Olvert A. Berkhemer and others},
   doi = {10.1056/nejmoa1411587},
   issn = {0028-4793},
   journal = {New England Journal of Medicine},
   pmid = {25517348},
   publisher = {Massachusetts Medical Society},
   title = {A Randomized Trial of Intraarterial Treatment for Acute Ischemic Stroke},
   year = {2015}
}

@inproceedings{Ha2018,
   author = {David Ha and Jürgen Schmidhuber},
   booktitle = {Advances in Neural Information Processing Systems},
   title = {Recurrent World Models Facilitate Policy Evolution},
   year = {2018}
}

@article{Karstensen2025,
   author = {Lennart Karstensen and others},
   doi = {10.1016/j.compbiomed.2025.110844},
   issn = {00104825},
   journal = {Computers in Biology and Medicine},
   month = {9},
   title = {Learning-based autonomous navigation, benchmark environments and simulation framework for endovascular interventions},
   volume = {196},
   year = {2025}
}

@article{Moosa2025,
   author = {Farhana Moosa and Harry Robertshaw and Lennart Karstensen and Thomas C. Booth and Alejandro Granados},
   doi = {10.1007/s11548-025-03360-x},
   issn = {18616429},
   issue = {6},
   journal = {Int J CARS},
   month = {6},
   pages = {1231-1238},
   pmid = {40299264},
   publisher = {Springer Science and Business Media Deutschland GmbH},
   title = {Benchmarking reinforcement learning algorithms for autonomous mechanical thrombectomy},
   volume = {20},
   year = {2025}
}

@article{Hu2023,
   author = {Anthony Hu and others},
   month = {9},
   title = {GAIA-1: A Generative World Model for Autonomous Driving},
   url = {http://arxiv.org/abs/2309.17080},
   year = {2023}
}

@inproceedings{Hansen2024,
   author = {Nicklas Hansen and Hao Su and Xiaolong Wang},
   booktitle = {The Twelfth International Conference on Learning Representations},
   title = {{TD-MPC2}: Scalable, Robust World Models for Continuous Control},
   year = {2024}
}

@InProceedings{Robertshaw3_2025,
    author = {Harry Robertshaw and Han-Ru Wu and Alejandro Granados and Thomas C. Booth},
    title = {World Model for {AI} Autonomous Navigation in Mechanical Thrombectomy},
    booktitle = {proceedings of Medical Image Computing and Computer Assisted Intervention -- MICCAI 2025},
    year = {2025},
    page = {683 -- 693}
}

@article{Fedorov2012,
   author = {Andriy Fedorov and others},
   doi = {10.1016/j.mri.2012.05.001},
   issn = {0730725X},
   journal = {Magnetic Resonance Imaging},
   pmid = {22770690},
   title = {3D Slicer as an image computing platform for the Quantitative Imaging Network},
   year = {2012},
}

@inproceedings{Sadati2025,
    author = {Hadi Sadati and others},
    title = {Autonomous Endovascular Navigation with a Long CTR: In-Vitro Studies},
    booktitle = {Hamlyn Symposium on Medical Robotics},
    year = {2025}
}

@inproceedings{Eyberg2022,
   author = {Christoph Eyberg and Lennart Karstensen and Tim Pusch and Johannes Horsch and Jens Langejürgen},
   doi = {10.1515/cdbme-2022-0023},
   issn = {23645504},
   issue = {1},
   booktitle = {Current Directions in Biomedical Engineering},
   pages = {89-92},
   title = {A ROS2-based Testbed Environment for Endovascular Robotic Systems},
   volume = {8},
   year = {2022}
}

@article{Karstensen2022,
   author = {Lennart Karstensen and others},
   doi = {10.1007/s11548-022-02646-8},
   issn = {18616429},
   issue = {11},
   journal = {Int J CARS},
   month = {11},
   pages = {2033-2040},
   pmid = {35604490},
   publisher = {Springer Science and Business Media Deutschland GmbH},
   title = {Learning-based autonomous vascular guidewire navigation without human demonstration in the venous system of a porcine liver},
   volume = {17},
   year = {2022},
}

@inbook{Wei2012,
   author = {Yiyi Wei and Stephane Cotin and Jrmie Dequidt and Christian Duriez and Jrmie Allard and Erwan Kerrie},
   doi = {10.5772/48635},
   journal = {Aneurysm},
   month = {8},
   publisher = {InTech},
   title = {A (Near) Real-Time Simulation Method of Aneurysm Coil Embolization},
   year = {2012},
}

@book{Faure2012,
   author = {François Faure and others},
   city = {Berlin, Heidelberg},
   isbn = {978-3-642-29014-5},
   pages = {283-321},
   publisher = {Springer Berlin Heidelberg},
   title = {SOFA, a Multi-Model Framework for Interactive Physical Simulation},
   year = {2012},
}

@article{Sun2013,
   author = {Chung Huan J. Sun and others},
   doi = {10.1161/CIRCULATIONAHA.112.000506},
   issn = {00097322},
   issue = {10},
   journal = {Circulation},
   month = {3},
   pages = {1139-1148},
   pmid = {23393011},
   title = {Picture to puncture: A novel time metric to enhance outcomes in patients transferred for endovascular reperfusion in acute ischemic stroke},
   volume = {127},
   year = {2013}
}

@article{Madden2004,
   author = {Michael G Madden and Tom Howley},
   doi = {https://doi.org/10.1023/B:AIRE.0000036264.95672.64},
   pages = {375–398},
   title = {Transfer of Experience between Reinforcement Learning Environments with Progressive Difficulty},
   journal = {Artificial Intelligence Review},
   volume = {21},
   year = {2004}
}

@inproceedings{Yin2017,
   author = {Haiyan Yin and Sinno Jialin Pan},
   doi = {https://doi.org/10.1609/aaai.v31i1.10733},
   booktitle = {Thirty-First AAAI Conference on Artificial Intelligence (AAAI-17)},
   title = {Knowledge Transfer for Deep Reinforcement Learning with Hierarchical Experience Replay},
   year = {2017}
}

@article{Jackson2023,
   author = {Benjamin Jackson and others},
   doi = {10.1007/s11548-023-02991-2},
   issn = {18616429},
   journal = {Int J CARS},
   title = {Comparative verification of control methodology for robotic interventional neuroradiology procedures},
   year = {2023},
}

@ARTICLE{Robertshaw2026ral,
  author={Robertshaw, H. and others},
  journal={IEEE Robotics and Automation Letters}, 
  title={Toward AI Autonomous Navigation for Mechanical Thrombectomy Using Hierarchical Modular Multi-Agent Reinforcement Learning ({HM-MARL})}, 
  year={2026},
  volume={11},
  number={4},
  doi={10.1109/LRA.2026.3664661}}

@article{Robertshaw2026jaha,
author = {Harry Robertshaw and others},
title = {A Position Statement on Endovascular Models and Effectiveness Metrics for Mechanical Thrombectomy Navigation, on Behalf of the Stakeholder Taskforce for Artificial Intelligence Assisted Robotic Thrombectomy ({START})},
journal = {Journal of the American Heart Association},
volume = {15},
year = {2026},
doi = {10.1161/JAHA.125.044931},
}

@misc{robertshaw2026iros,
      title={Toward Safe Autonomous Robotic Endovascular Interventions using World Models}, 
      author={Harry Robertshaw and others},
      year={2026},
      eprint={2604.20151},
      archivePrefix={arXiv},
      primaryClass={cs.RO},
      url={https://arxiv.org/abs/2604.20151}, 
}

@misc{williams2015,
      title={Model Predictive Path Integral Control using Covariance Variable Importance Sampling}, 
      author={Grady Williams and Andrew Aldrich and Evangelos Theodorou},
      year={2015},
      eprint={1509.01149},
      archivePrefix={arXiv},
      primaryClass={eess.SY},
      url={https://arxiv.org/abs/1509.01149}, 
}

\vfill

\end{document}